\documentclass[11pt]{article}

\usepackage[preprint]{acl}

\usepackage{times}
\usepackage{latexsym}
\usepackage{enumitem}

\usepackage[T1]{fontenc}

\usepackage[utf8]{inputenc}

\usepackage{microtype}

\usepackage{inconsolata}

\usepackage{graphicx}

\usepackage{fontawesome5}

\usepackage{xurl}
\usepackage{subcaption}
\usepackage{linguex}
\usepackage{amsmath,amssymb}
\usepackage[capitalize,noabbrev]{cleveref}
\usepackage{xspace}
\usepackage{booktabs}
\usepackage{tcolorbox}
\tcbuselibrary{listings}
\usepackage[dvipsnames]{xcolor}
\definecolor{cudOrange}{HTML}{D55E00}
\definecolor{cudGreen}{HTML}{03AF7A}
\definecolor{cudPink}{HTML}{CC79A7}
\definecolor{cudBlue}{HTML}{005A9C}
\definecolor{cudGray}{HTML}{999999}
\definecolor{cudVermilion}{HTML}{D55E00}
\definecolor{cudBrightOrange}{HTML}{E69F00}
\definecolor{cudBrightBlue}{HTML}{0072B2}
\definecolor{cudSkyBlue}{HTML}{56B4E9}

\newcommand{\gptsmall}{\texttt{Small}\xspace}
\newcommand{\gptmedium}{\texttt{Medium}\xspace}
\newcommand{\gptlarge}{\texttt{Large}\xspace}
\newcommand{\gptxl}{\texttt{XL}\xspace}

\newcommand{\baseline}{\textcolor{cudBlue}{\textsc{Singleton-Disrupted}}\xspace}
\newcommand{\linking}{\textcolor{cudVermilion}{\textsc{Coref-Disrupted}}\xspace}

\newcommand{\dll}{\ensuremath{\Delta_{\mathit{LL}}}\xspace}
\newcommand{\ddll}{\ensuremath{\Delta \dll}\xspace}

\usepackage{etoolbox}
\AtBeginDocument{
  \setlength{\abovedisplayskip}{5pt}
  \setlength{\belowdisplayskip}{5pt}
  \setlength{\abovedisplayshortskip}{0pt}
  \setlength{\belowdisplayshortskip}{3pt}
}

\title{
    Using LMs to Model the Effects of Context and Coreference during Sentence Comprehension
}

\author{Kohei Kajikawa$^{*1}$ \quad Lin Ai$^{*1}$ \quad Tatsuki Kuribayashi$^{2,3}$ \quad Ethan Gotlieb Wilcox$^1$ \\
$^1$Department of Linguistics, Georgetown University, USA \\
$^2$Computing and Mathematical Sciences Division, MBZUAI, UAE \\
$^3$Center for Language AI Research, Tohoku University, Japan \\
{\small
    \textbf{Correspondence:} \href{mailto:kk1571@georgetown.edu}{\texttt{kk1571@georgetown.edu}},
    \href{mailto:la987@georgetown.edu}{\texttt{la987@georgetown.edu}}}\\
{\small * authors contributed equally}
}

\begin{document}
\maketitle

\begin{abstract}
	Language models (LMs) are often used as a tool to model human language processing. Recent studies suggest that severely restricting LMs' context window improves their fit to human psycholinguistic data by simulating human working memory constraints. However, it is possible that this strict memory-decay approach overlooks humans' reliance on long-range structural representations, such as discourse structre. In this work, we systematically vary the context window size of GPT-2 across four large-scale naturalistic English reading-time datasets and observe a U-shaped relationship: Although restricted contexts ($<$ 20 tokens) successfully capture local memory limitations, expanded contexts (500--1,000 tokens) ultimately yield the highest overall psycholinguistic fit. To investigate the mechanism driving this benefit, we conduct a \emph{counterfactual} inference-time experiment that disrupts cross-sentential entity chains by pronominalizing repeated discourse entities. Obscuring these structural linkages significantly degrades the predictive power of larger context windows by 20\% to 40\%. Our experiments demonstrate that tracking long-range coreference relations is one important factor for the alignment between LM surprisal and human reading behavior, and approximate the extent to which human comprehenders use global discourse relations during language processing.\footnote{Our code is available at \faGithub~\url{https://github.com/kohei-kaji/contextsize}.}
\end{abstract}

\section{Introduction}

How do people integrate past information to predict what they will encounter next when reading or listening?
In recent years, language models (LMs) have emerged not merely as engineering achievements~\citep{radford-etal-2019,brown-etal-2020}, but as useful cognitive models to investigate human psycholinguistic mechanisms~\citep{goldstein-etal-2022,frank-goodman-2026}.
LMs' surprisal values have been demonstrated to be well correlated to human reading times~\citep{wilcox-etal-2020,oh-etal-2022,kuribayashi-etal-2024,kuribayashi-etal-2025}.
However, recent scaling of LMs has also revealed a noteworthy divergence: as models grow larger and achieve lower perplexity, their surprisal estimates often become less predictive of human reading behavior~\citep{oh-schuler-2023,shain-etal-2024}.
One leading hypothesis suggests this discrepancy arises because modern LMs are effectively ``superhuman''~\citep{oh-linzen-2026} in their prediction capabilities, possessing a long tail of factual knowledge and lossless short-term memory that bypass the cognitive constraints inherent to human language processing.

To better align model predictions with human reading behavior, recent computational investigations have attempted to restrict models' access to prior context, simulating human working memory limits~\citep{hahn-etal-2022-resource,kuribayashi-etal-2022,timkey-linzen-2023,de-varda-marelli-2024,clark-etal-2025, xu-etal-2026}.
Studies have shown that constraining the context window size~\citep{kuribayashi-etal-2022} or introducing a linear recency bias to the attention mechanism~\citep{de-varda-marelli-2024,clark-etal-2025} significantly improves the fit of LM surprisal to naturalistic human reading times.
Yet, this strict memory-decay approach introduces a theoretical gap. 
It overlooks how people actively build and maintain structured discourse representations over extended narratives~\citep{karttunen-1969-discourse-referents,kintsch-vandijk-1978,grosz-etal-1995,ericsson-kintsch-1995,kintsch-1998,jaffe-etal-2018,tsipidi-etal-2024}.
For instance, reading times have been shown to be highly sensitive to long-range contextual information, such as the number of possible coreferences \citep{jaffe-etal-2018}.
Furthermore, the distribution of information across a text is not merely random, but rather fluctuates systematically according to hierarchical discourse structures~\citep{tsipidi-etal-2024, tsipidi-etal-2025-harmonic}.
This suggests that comprehenders actively use long-range discourse linkages to drive forward-looking predictions.
Such capacities have been treated theoretically, for example, by the \emph{long-term working memory} framework~\citep{ericsson-kintsch-1995,kintsch-1998}, which posits skilled cognitive mechanisms for maintaining extended narrative states.
Our hypothesis is that models that cannot take advantage of such discourse linkages will be insufficient as models of human language processing.

We investigate how context length and discourse structure influence the cognitive plausibility of language models: we systematically vary the context window size of GPT-2 and evaluate its predictive power for human reading time in English.
We observe a U-shaped relationship between context size and psychometric predictive power.
At short time-scales ($<$ 20 tokens), restricted context provides a better fit to human data compared to longer one, successfully capturing localized working memory constraints observed by~\citet{kuribayashi-etal-2022}.
Crucially, however, at larger time scales ($>$ 20 tokens), expanded contexts become highly beneficial once again, with predictive power peaking at extended lengths of roughly 500 to 1,000 tokens.
This U-shaped curve suggests that reading times reflect the interplay of two factors: local working memory constraints and global discourse integration.

Second, we ask \emph{why} this extended context facilitates the predictive power of surprisal. 
We hypothesize that larger context windows are beneficial not merely because they provide a larger \emph{bag of words}, but because coreference relations beyond sentence boundaries help maintain discourse coherence and enable people to predict what is coming next~\citep{hobbs1979coherence}.
While it is known that readers with greater working memory capacity are more successful at resolving pronouns~\citep{daneman-carpenter-1980}, the predictive benefits of an extended context window extend beyond simple pronominal anaphora.
For example, consider the sentence pair: \textit{Mary pushed Sue. The poor girl \dots}
Most comprehenders will establish a coreference link between \textit{the poor girl} and \textit{Sue}, and doing so dramatically reduces the entropy over possible next-word continuations.
Similarly, Transformer-based LMs have been shown to establish coreference relations between noun phrases with or without lexical overlap \cite{tenney-etal-2019-bert,clark-etal-2019-bert,sorodoc-etal-2020-probing}.
Therefore, we posit that LMs' ability to represent coreference cues partially accounts for their better alignment with human reading times as context size increases. 

To test this, we examine LMs' predictions for reading times on \emph{counterfactual} versions of the test materials where coreference relations are obscured.
We analyze to what extent this degrades the models' ability to predict the reading times of people who have full knowledge of entity relationships.
We find that when entity coreferences are pronominalized, the advantage in predictive power of larger context windows drops by approximately 20\% to 40\%, substantially above a baseline manipulation.
This drop also provides a quantitative measure of the importance of coreference resolution in discourse processing. 
More broadly, our approach of reading-time modeling with a counterfactual context provides a generalizable framework for evaluating how specific aspects of context drive expectation-based human language comprehension.

\section{Evaluating Surprisal on Reading Times}\label{sec:method}

We adopt surprisal theory \citep{hale-2001, levy-2008, smith-levy-2013} as a framework for linking LMs' outputs and human psycholinguistic processing measures.
Building on this framework, we investigate the relationship between LMs' context size and discourse structure.

\subsection{Reading-time Datasets}\label{sec:datasets}
We use four large-scale English reading-time datasets: Brown~\citep{smith-levy-2013}, Natural Stories~\citep{futrell-etal-2021}, OneStop~\citep{berzak-etal-2025}, and Provo~\citep{luke-christianson-2018}.\footnote{
    Although \citet{kuribayashi-etal-2022} evaluated their models on the Dundee corpus~\citep{kennedy-etal-2003}, we exclude this dataset from our study because it features only 10 participants and is not publicly available.
}
Brown consists of 13 passages (7,234 words) of \textbf{self-paced reading} (SPR) data from 35 native English speakers.
Natural Stories consists of 10 naturalistic narratives (10,256 words).
We use SPR data from 181 native English speakers~\citep{futrell-etal-2021} and \textbf{A-Maze} reading times from 95 native English speakers~\citep{boyce-levy-2023}.
OneStop consists of 10 articles (35,181 words) with eye-tracking data from 180 native English speakers.
We use the ``ordinary reading'' sub-portion.
Provo consists of 55 paragraphs (2,745 words) of eye-tracking data from 84 native English speakers.
\Cref{tab:token_stats} summarizes the descriptive statistics of document length in GPT-2 tokens for each dataset.

In SPR, participants read texts word-by-word by pressing a button, with the interval between presses recorded as the reading time.
In A-Maze, a variant of the Maze task~\citep{forster2009maze}, participants sequentially choose the correct next word from two options, and this choice time is taken as the reading time.
For the eye-tracking datasets, we analyze three eye-tracking measures: 
\textbf{first fixation} (FF), the duration of initial fixation to a region during the first pass; \textbf{gaze duration} (GD), which sums all fixations on a region before the reader's gaze leaves the region in any direction; and \textbf{total fixation} times (TF), the sum of all fixations on a region.

We apply several preprocessing steps.
For all datasets, we exclude the first and last words of each sentence and any words containing punctuation.
For SPR data (Brown and Natural Stories), we follow \citet{futrell-etal-2021} and exclude reading times outside the 100--3,000 ms range.
We also exclude data from participants with poor comprehension, specifically those scoring less than $5/6$ on the comprehension questions for NS SPR and under 80\% accuracy on A-Maze~\citep{boyce-levy-2023}.
For eye-tracking data, we remove skipped words.
For all datasets, we model the mean processing time across participants~\citep{smith-levy-2013,wilcox-etal-2023-testing} to isolate item-level variance and ensure computational feasibility, though we acknowledge the inability to model subject-level random effects as a limitation.

\begin{table}[t]
    \centering
    \setlength{\tabcolsep}{4pt}
    \begin{tabular}{lrrr}
        \toprule
        \textbf{Dataset} & \textbf{Mean} & \textbf{SD} & \textbf{Range}\\
        \midrule
        Brown           & 707.00  & 199.59 & 313--959\\
        Natural Stories & 1237.30 & 68.28 & 1156--1359 \\
        OneStop         & 751.27  & 171.57 & 450--1271\\
        Provo           & 59.53   & 8.20 & 43--80\\
        \bottomrule
    \end{tabular}
    \caption{Descriptive statistics (mean, standard deviation, and range) of the number of tokens by the GPT-2 tokenizer per document across datasets. Provo is notably shorter than the others.}
    \label{tab:token_stats}
\end{table}

\subsection{Surprisal Estimates}\label{sec:surp}
We use four pretrained GPT-2 models (\gptsmall, \gptmedium, \gptlarge, and \gptxl),\footnote{We use the Transformers library~\citep{wolf-etal-2020-transformers}} as prior work has shown that GPT-2 surprisal, particularly by the \gptsmall variant, is the best predictor of English reading times compared to larger language models~\citep{oh-schuler-2023,shain-etal-2024}.
To estimate surprisal, we adopt the whitespace-trailing decoding~\citep{oh-schuler-2024}, which reassigns the probability of a leading whitespace to the preceding word.
While GPT-2 utilizes at most 1,024 tokens to predict the next token, the use of this decoding strategy restricts our maximum context window to 1,023 tokens.
We use 28 different context lengths (in subword tokens) from 1 to 1,023 to estimate surprisal for each word.\footnote{The context sizes adopted in this study: [1, 2, 3, 4, 5, 6, 7, 10, 12, 16, 20, 25, 32, 40, 50, 64, 80, 100, 128, 160, 200, 256, 320, 400, 512, 640, 800, 1023].}
Following \citet{kuribayashi-etal-2022}, the BOS token is not included in the context as it can cause LMs to misinterpret the beginning of the sentence.
The mean surprisal values for each dataset across all context sizes are provided in \cref{fig:meansurp} in \cref{app:meansurp}.

\subsection{Regression Analysis}\label{sec:stat}

To quantify the predictive power of LM-based surprisal on reading time, we follow the methodology established in prior psycholinguistic studies~\citep[e.g.,][]{frank-bod-2011,goodkind-bicknell-2018,wilcox-etal-2020,wilcox-etal-2023-testing} and measure the improvement in regression model fit afforded by surprisal (\dll).
Specifically, we compare the log-likelihood of a baseline regression model (excluding surprisal) against that of a target regression model (including surprisal) on held-out test data.

More precisely, the dataset $\mathcal{D}=\{(y_t,\mathbf{x}_t)\}^N_{t=1}$ is defined such that $y_t$ represents the mean reading time for word $w_t$, $\mathbf{x}_t$ is a vector of predictors for $w_t$, and $N$ is the total number of data points.
We fit a linear regression model of the following form:
\begin{align}
    y_{t} = \beta_0 + \mathbf{x}_{t}^\top \boldsymbol{\beta} + \varepsilon_{t}, \quad \varepsilon_{t} \overset{\text{iid}}{\sim} \mathcal{N}(0, \sigma^2),  \label{eq:regression}
\end{align}
where $\beta_0$ is the intercept and $\boldsymbol{\beta}$ is the vector of coefficients.
The predictors $\mathbf{x}_t$ in the baseline model include word position within a document, word length (number of characters), and unigram surprisal estimated from about 33 billion tokens of the Pile dataset \citep{gao-etal-2020}.
We also incorporate two spillover positions (i.e., the values at $t-1$ and $t-2$) for both word length and unigram surprisal.
The target model is constructed by adding surprisal of the current word, along with its two spillover positions, to the baseline model.
All predictors are $z$-scored.

The predictive power of surprisal is evaluated with 10-fold cross-validation.
The model parameters are estimated on the nine folds and evaluated on the held-out fold.
For every data point $t$ in the held-out fold, we compute the difference in log-likelihood between the target and baseline models:
\begin{align}
     \Delta_{\mathit{LL}}^{(t)} = \ell_{\text{target}}^{(t)} - \ell_{\text{baseline}}^{(t)},
\end{align}
where $\ell_{\text{model}}^{(t)}$ is the log-likelihood for a single observation at $t$ under the normal distribution defined in \cref{eq:regression}.
The final \dll metric is calculated as the mean $\Delta_{\mathit{LL}}^{(t)}$ across data points in $\mathcal{D}$.

Thus, \dll quantifies the extent to which LM surprisal accounts for reading-time variance beyond what is already explained by the baseline model.

\section{Experiment 1: The Effect of Context Window Size}\label{sec:exp1}

To investigate how an LM's context size influences the alignment between its surprisal and human reading times, we systematically vary the number of context tokens provided to GPT-2 models (\cref{sec:surp}).
We then evaluate the predictive power of the resulting surprisal by using the method in~\cref{sec:stat} across the four large-scale English reading-time datasets described in~\cref{sec:datasets}.\footnote{
    As one of the reviewers suggested, we also evaluated models other than GPT-2 on the Natural Stories SPR.
    The results, presented in \Cref{app:ns_biggermodels}, show that our overall findings remain consistent across different models.
}

\subsection{Results}
\begin{figure*}[t]
    \centering
    \includegraphics[width=\linewidth]{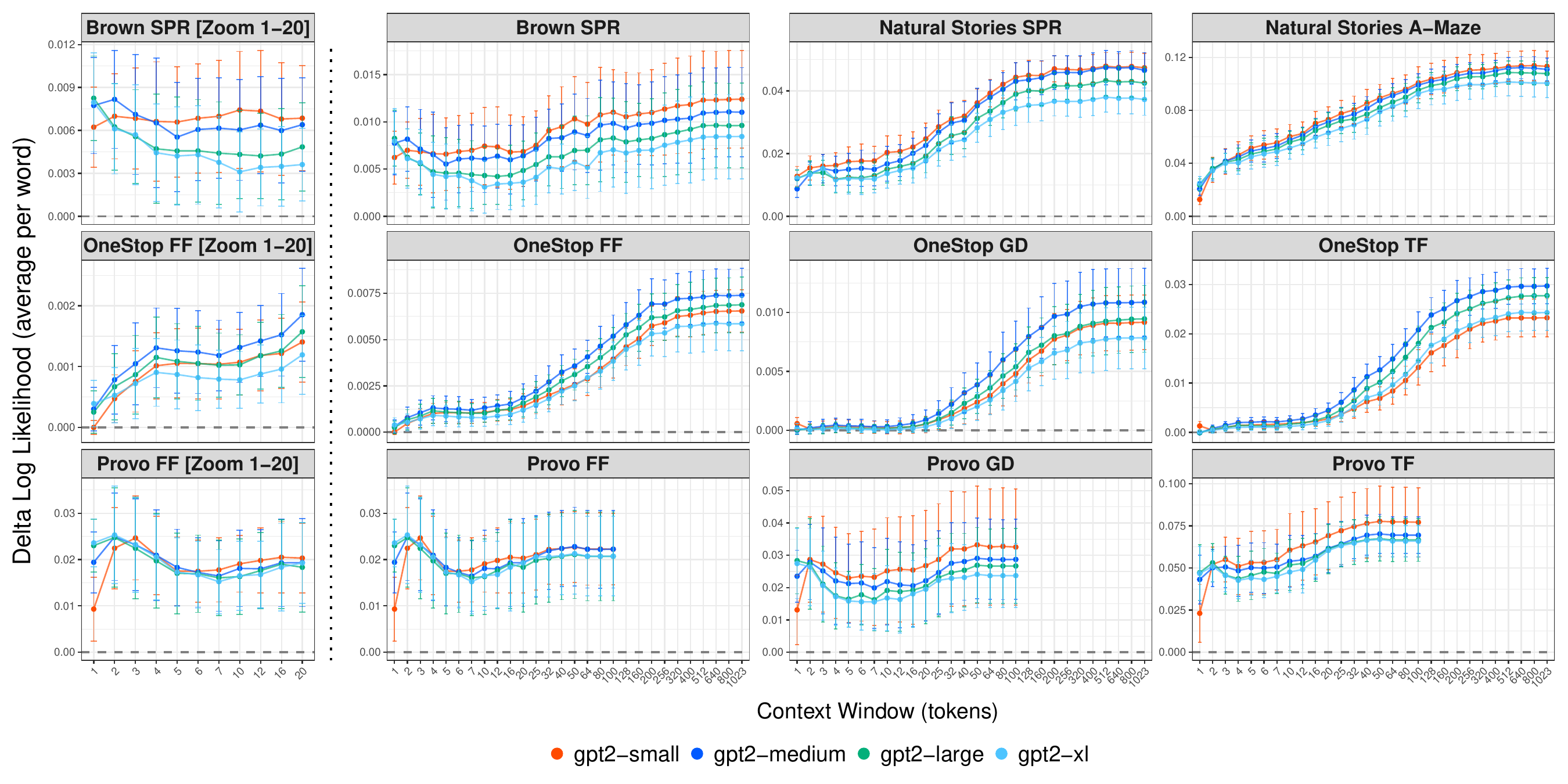}
    \caption{Predictive power (\dll) across different context window sizes. Each point represents the \dll with 95\% confidence interval.
    For the Provo dataset, results are shown up to the document's maximum length.
    The leftmost column displays zoomed-in plots for Brown, OneStop FF, and Provo FF, focusing on context windows of 1--20 tokens.
    Abbreviations: SPR = self-paced reading; FF = first fixation; GD = gaze duration; TF = total fixation time.
    }
    \label{fig:dll}
\end{figure*}

The relationship between context size and the fit of surprisal to reading times is illustrated in~\cref{fig:dll}.
Looking at the macro trend, except for Provo FF, we observe that expanding the context window size enhances predictive power across datasets and model variants.
This trend is robust in datasets featuring longer documents (Brown, Natural Stories, and OneStop), whereas it is less consistent in Provo, which consists of shorter documents (see also~\cref{tab:token_stats}).
To test this trend empirically, we fit a Bayesian mixed-effects model using \texttt{brms}~\citep{brms} in R~\citep{r-2025} with \dll as the response, log-transformed context length as a fixed effect, and model and dataset as random intercepts.
We observed a robust positive effect of context length ($\hat{\beta} = 4.53 \times 10^{-3}$, 95\% CrI: $[4.23, 4.83] \times 10^{-3}$).\footnote{
We assigned weakly informative priors to the fixed effect ($\mathcal{N}(0, 0.1)$ and variance components ($\operatorname{Exponential}(1)$). The model was estimated using 4 Markov chains with 2,000 iterations per chain, including 1,000 warmup iterations. All parameters successfully converged ($\hat{R} < 1.01$).
}

However, a closer inspection of shorter context windows (up to roughly 20 tokens) reveals a more nuanced pattern.
With the exception of Natural Stories A-Maze, we observe a U-, or sometimes S-shaped pattern.
This U-shape is especially pronounced in larger models and in measures that reflect earlier stages of cognitive processing, such as first fixation (FF) in eye-tracking.
Within this short-context regime, the \emph{longer-is-better} relationship does not strictly hold.
Instead, local optima emerge at context sizes of approximately 2 to 4 tokens, though the exact peak varies depending on the model scale and dataset.
This finding is consistent with \citet{kuribayashi-etal-2022}, who demonstrated that restricting context to within sentence boundaries can yield a better fit to human reading data.

These results suggest a dual mechanism of context effects: within the short-context region, a \emph{shorter-is-better} trend operates locally, indicating that reading times are highly sensitive to immediate, local statistics.
Conversely, as the model is granted access to increasingly longer cross-sentential context, the \emph{longer-is-better} trend emerges, ultimately yielding the highest overall fit.

Finally, although tangential to our primary investigation of context effects, we observed a notable result regarding model scale on OneStop, a dataset released in 2025 with the largest number of participants~\citep{berzak-etal-2025}: surprisal from GPT2-\gptmedium consistently outperforms that from GPT2-\gptsmall as a predictor of reading times.
This deviates from the prevailing consensus in recent literature, which has identified surprisal from GPT2-\gptsmall as the best predictor of English reading times~\citep{oh-schuler-2023,shain-etal-2024}.

\subsection{Follow-up Analysis: Localization of the Context Effect}\label{sec:followup}

While Experiment 1 demonstrated that extended contexts yield the highest overall fit, the underlying mechanism remains unclear.
To investigate, we analyze the relationship between part-of-speech (POS) tags and word-by-word variations in both raw surprisal and model fit.
We hypothesize that if long contexts are useful specifically because they help to accurately model discourse state, then the predictive gains should fall on words that are important for establishing discourse relations (e.g., coreference relations and predicate-argument structures;~\citealp{hobbs1979coherence,grosz-etal-1995}), such as nouns and predicates, but not on determiners and particles.

We use Stanza~\citep{qi-etal-2020} to assign Universal Dependencies POS tags~\citep{de-marneffe-etal-2021} to each word. To ensure statistical reliability, our analysis is restricted to major syntactic categories with at least 100 occurrences across the datasets. The evaluated tags include open class words (\texttt{ADJ}, \texttt{ADV}, \texttt{NOUN}, \texttt{PROPN}, \texttt{VERB}) and closed class words (\texttt{ADP}, \texttt{AUX}, \texttt{CCONJ}, \texttt{DET},  \texttt{NUM}, \texttt{PART}, \texttt{PRON}, \texttt{SCONJ}). 

To isolate the localized advantage of long-range context, we compare the \gptsmall estimates under a \emph{short} context baseline (2 tokens) against a \emph{long} context setup (up to 1,023 tokens or the maximum length permitted by the document).
For each word position $t$, we evaluate two metrics.
First, to localize and quantify the shifts in surprisal values induced by an expanded context window, we define the delta surprisal as:
\begin{align}
    \Delta\operatorname{surp}^{(t)} = \operatorname{surp}_{\text{Short}}(t) - \operatorname{surp}_{\text{Long}}(t).
\end{align}
Second, to capture the incremental improvement in predictive power for reading time, we compute the pointwise change in log-likelihood between the long and short context regression models:
\begin{align}
    \Delta\dll^{(t)} = \ell_{\text{Long}}^{(t)} - \ell_{\text{Short}}^{(t)},
\end{align}
where $\ell_{\text{context}}^{(t)}$ denotes the log-likelihood at $t$ under the respective context size condition.

To assess these localized effects while accounting for dataset-specific variance, we fit Bayesian mixed-effects models with \texttt{brms}.
We model both $\Delta\operatorname{surp}^{(t)}$ and $\Delta\dll^{(t)}$ as functions of the categorical POS predictors, and include random slopes for the respective datasets. 

\subsubsection{Results}
\begin{figure}[t]
    \centering
    \begin{subfigure}[b]{\linewidth}
        \centering
        \includegraphics[width=0.8\linewidth]{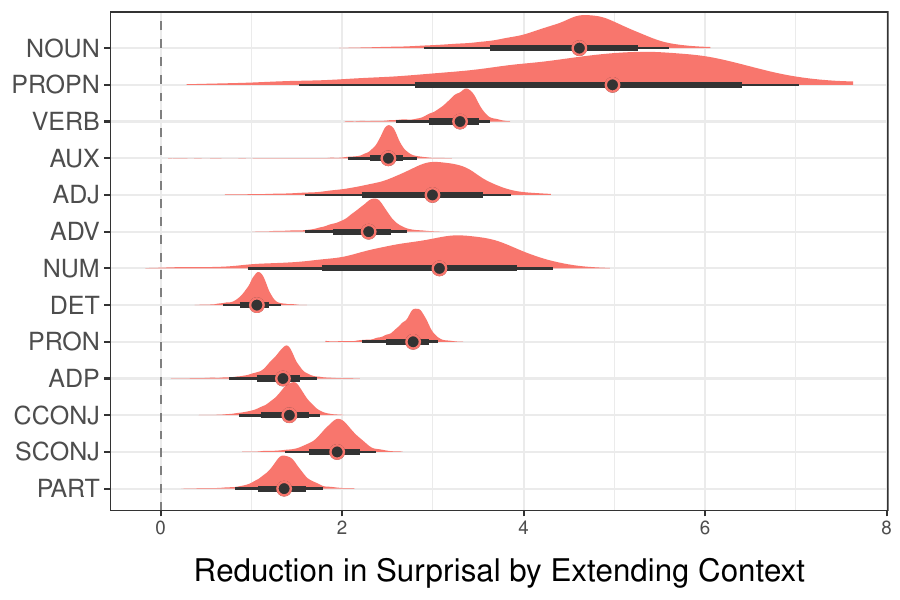}
        \caption{Posterior estimates of $\Delta\operatorname{surp}^{(t)}$ across POS tags. Positive values indicate that long-range context systematically reduces the model's surprisal for that category.}
        \label{fig:pos_surp}
    \end{subfigure}
    \vspace{0.5em}
    \begin{subfigure}[b]{\linewidth}
        \centering
        \includegraphics[width=0.8\linewidth]{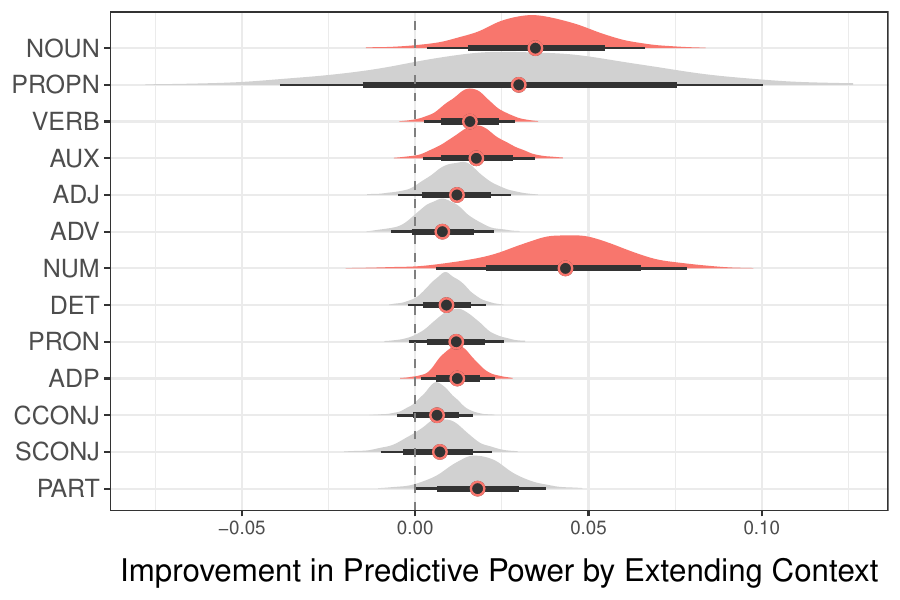}
        \caption{Posterior estimates of $\Delta\dll^{(t)}$ across POS tags. Positive values indicate a greater reading-time predictive benefit from an extended context window at that category.}
        \label{fig:pos_ddll}
    \end{subfigure}
    \caption{Localization analysis of context effects across POS tags using Bayesian mixed-effects models.
    Densities show the central 99\% of the posterior distributions. 
    Points represent posterior medians, while the thick and thin horizontal bars denote the 80\% and 95\% credible intervals, respectively.
    Red distributions indicate that the 95\% credible interval does not contain zero.}
    \label{fig:pos_analysis}
\end{figure}

The posterior estimates reveal a divergence between the reduction in model surprisal and the improvements in reading-time prediction.\footnote{
We assigned weakly informative priors to the fixed effects ($\mathcal{N}(0, 2)$ for $\Delta\operatorname{surp}^{(t)}$ and $\mathcal{N}(0, 0.1)$ for $\Delta\dll^{(t)}$) and variance components ($\operatorname{Exponential}(1)$). Both models were estimated using 4 Markov chains with 2,000 iterations per chain, including 1,000 warmup iterations. All parameters successfully converged ($\hat{R} < 1.01$).
}
As shown in \cref{fig:pos_surp}, expanding the context window size systematically reduces surprisal across all evaluated POS tags.
This surprisal reduction is most pronounced for proper nouns (\texttt{PROPN}), nouns (\texttt{NOUN}), verbs (\texttt{VERB}), numerals (\texttt{NUM}), and adjectives (\texttt{ADJ}). 

However, these localized drops in surprisal do not uniformly translate to better psychometric fit (\cref{fig:pos_ddll}).
Most notably, while \texttt{PROPN} exhibits the largest surprisal reduction, its improvement in predictive power ($\Delta\dll^{(t)}$) is not statistically significant due to wide variance across datasets.
Instead, the statistically significant gains in $\Delta\dll^{(t)}$ are concentrated in \texttt{NOUN}, \texttt{NUM}, \texttt{VERB}, auxiliry (\texttt{AUX}), and adpositions (\texttt{ADP}).
That being said, the two biggest gains from larger contexts are assigned to words establishing discourse relations such as nominals and predicates, while determiners and particles, for example, show no significant gains from context, which is consistent with our original hypothesis.

These findings demonstrate that an extended context window does not merely act as a uniform mechanism for benefiting prediction.
Rather, the predictive gains are concentrated primarily on nominals and predicates.
We hypothesize that these categories benefit most because they serve to represent discourse entities and events.
This suggests that expanding the context window enables the model to implicitly track coreference relations and maintain the global discourse states.
We turn to this question in the next section.

\section{Experiment 2: Disrupting Coreference Relations during Inference}\label{sec:exp2}

An important aspect of language processing is figuring out who is doing what to whom. This type of information is supplied by referring expressions that provide links to discourse entities.

During comprehension, people incrementally update their mental model about the entities being referred to by tracking the coreference relations~\citep{karttunen-1969-discourse-referents}.
Coreference resolution, the task of establishing the identity between two referring expressions with or without the same wording, is also an important NLP task \citep{jm3}.
LMs have also been shown to establish coreference relations across long documents \citep{tenney-etal-2019-bert,clark-etal-2019-bert, sorodoc-etal-2020-probing}.
We hypothesize that it is specifically this ability that drives the alignment between long-context surprisal and human reading time.
We predict that when a model's ability to track entities is hindered, the advantage of long-context prediction will be degraded.

In order to test this prediction, we perform inference-time manipulations in which we pronominalize all entities that occur more than once, creating a \linking version of the test materials used in Experiment 1.
This removes information on the entity required to establish unambiguous coreference links with other mentions of the same entity in the context.
We then analyze to what extent this degrades the model's ability to predict reading times of people who have access to the full, discourse-linked context.
Decrease in reading time prediction indicates that the counterfactual context loses the structural information useful for simulating human language processing.

However, the pronominalization manipulation will significantly alter the text---for example, changing its overall entropy---and may lead to unnatural fluctuations in surprisal.
To account for these confounds, we create a \baseline version of the original text where we pronominalize all the entities that appear only once. 
This version of the text is pronominalized, but long-context dependency information is preserved; thus, reading time prediction should degrade only in the \linking condition (not in \baseline) if the coreference chain is important for simulating human reading.
Overall, 21,640 tokens are changed in \baseline and 22,695 tokens are changed in \linking. The number of tokens changed in \baseline and \linking are roughly equivalent, ensuring that the changes in delta log-likelihood do not just result from pronominalization but from the loss of long-context dependencies.

To illustrate our two conditions, consider the examples below adapted from Brown, where \ref{bluethroat-baseline} is \baseline and \ref{bluethroat-linking} is \linking.
Pronominalized text is colored.
Version \ref{bluethroat-baseline} contains several local infelicities, but maintains information about the main entities of the narrative.
Version \ref{bluethroat-linking} is perfectly fluent, but makes all references to entities ambiguous.

\begin{figure*}[t]
    \centering
    \includegraphics[width=\linewidth]{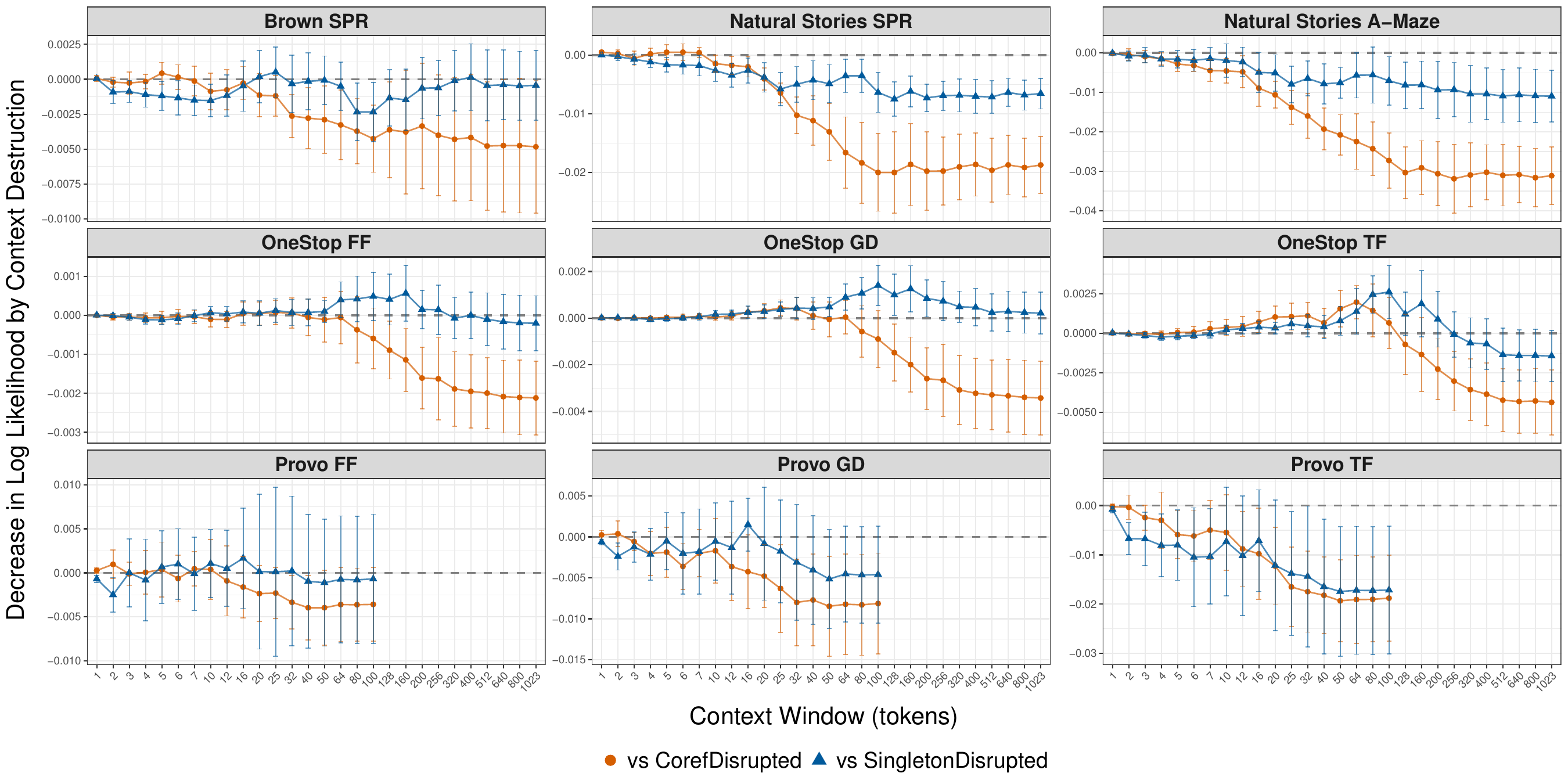}
    \caption{Difference of predictive power (\ddll) between Experiment 1 and our two target conditions---\baseline and \linking. Results are for \gptsmall.}
    \label{fig:delta_gpt2}
\end{figure*}

\ex. \label{bluethroat-baseline}
\textcolor{cudBlue}{They} hastily vacated \textcolor{cudBlue}{them} as the tall buffalo hunter pushed open \textcolor{cudBlue}{them} and walked towards \textcolor{cudBlue}{it}. Only Blue Throat stayed where he was, \textcolor{cudBlue}{his} supporting \textcolor{cudBlue}{his}. He leered at the stranger.

\ex. \label{bluethroat-linking}
Customers hastily vacated their tables as \textcolor{cudVermilion}{he} pushed open the doors and walked towards the bar. Only \textcolor{cudVermilion}{he} stayed where he was, his elbows supporting his massive frame. He leered at \textcolor{cudVermilion}{him}.

Both texts are out of distribution for an LM trained on naturally occurring English.
We predict that, even though texts like \ref{bluethroat-linking} may be perceived as more fluent, conditioning on them will yield surprisal estimates with lower predictive power for reading times, especially when the context window is long.

\subsection{Methods}

\subsubsection{Dataset Creation.}
Our methods are largely the same as those described in Section \ref{sec:method}, except for the datasets on which surprisal values are conditioned.
To create these datasets, we use the coreference parser from Stanza~\citep{qi-etal-2020,liu-etal-2024-mscaw}\footnote{The accuracy of this model is reported to be 95.7\% on the CorefUD dataset~\citep{nedoluzhko-etal-2022-corefud}.} to tag entities in our test materials.
We use silver entity tags to identify singleton and coreferent entities, which are replaced in the \baseline and \linking conditions, respectively.
We use the following methods to replace entities:
All tokens associated with an entity are replaced by a \texttt{[mask]} token.
We do not mask out entities if part of the entity falls outside of a model's context window.
We then prompt \mbox{DeepSeek-V4-Flash}~\citep{deepseekai2026deepseekv4} to collapse each contiguous \texttt{[mask]} span into a single appropriate pronoun, providing the original sentence and the masked sentence as context.
If no pronoun is produced, we use \textit{it} as a fallback.
For the full prompt template, see \cref{app:prompt}.
We do not change entities that are already pronominalized.

This procedure results in three parallel texts.
However, the \baseline and \linking conditions are shorter, as multi-token entities are replaced with a single-token pronoun.
To account for this, we fill extra tokens from the previous context chunk in the same condition.

\subsubsection{Text Comprehensibility Check}
To validate the naturalness of these two versions after perturbations, we evaluate their comprehensibility.
We conduct this evaluation on Natural Stories by having DeepSeek-V4-Flash answer reading comprehension questions based on the texts.
To avoid the risk of data contamination, where the LLM might have memorized the publicly available questions from the original datasets, we create a customized set of comprehension questions for Natural Stories inputs.
The questions are given in \Cref{app:comp_questions}.
Overall, DeepSeek-V4-Flash achieves 100\% (60/60) correct on the original text, 91.7\% (55/60) on \linking and 90\% (54/60) correct on \baseline.

\subsubsection{Surprisal Estimation}
We use the same GPT-2 models as in Experiment 1 to obtain surprisal values.
We then calculate \dll for each condition (\baseline and \linking) and subtract each from the \dll value obtained for the unaltered text in Experiment 1.
This metric is the same \ddll as in \Cref{sec:followup}, but calculated against the unaltered text for each condition, as opposed to short vs. long contexts in \Cref{sec:followup}.
When computing surprisal for entities as target tokens, we always un-pronominalize the entity first; the only difference between the conditions is therefore the context in which the surprisal is conditioned.

\subsection{Results}

Results for \gptsmall are shown in \cref{fig:delta_gpt2}, with \ddll plotted on the $y$-axis.
Results for all models are presented in \cref{app:delta_breakdown}, and are consistent across all models tested.

We observe two major trends.
First, when conditioning on local contexts ($\lessapprox$ 10 tokens), \ddll is very close to zero, indicating little change from the unaltered baseline.
In three cases (Brown SPR, Natural Stories SPR, and Provo TF), \baseline shows slightly negative \ddll, likely due to the fact that \baseline disrupts local semantic predictability.

However, as the context size gets larger, \linking shows a large decrease \ddll.
This is greater than the decrease observed in the \baseline condition in most cases, except for Provo.
To test this statistically, we conducted paired word-level two-tailed $t$-tests.
At the maximal context length, the \ddll in the \linking condition is significantly lower than zero across all datasets at $p<0.01$ except for Provo FF ($p=0.091$), and significantly less than the \ddll in the \baseline condition for Natural Stories SPR, Natural Stories A-Maze, One Stop FF, and One Stop GD ($p<0.001$ for all), as well as for Brown and One Stop TF ($p<0.05$ for both).
We do not observe differences between our two conditions in Provo, likely due to its short-length documents.

\begin{figure}[t]
    \centering
    \includegraphics[width=0.7\linewidth]{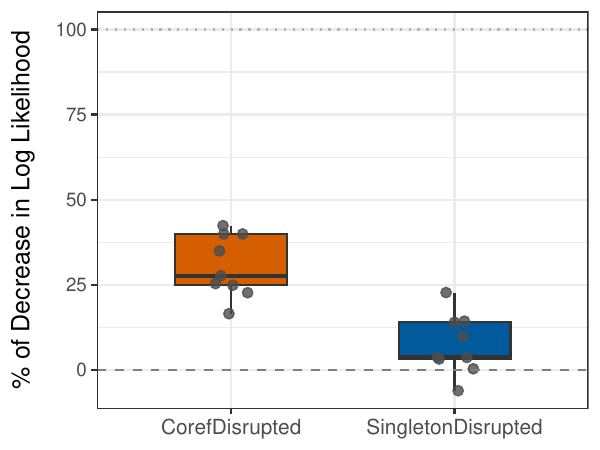}
    \caption{Distribution of the percentage decrease in log-likelihood under \linking and \baseline conditions for \gptsmall at a maximum context across all datasets.}
    \label{fig:percent-reduction}
\end{figure}

\cref{fig:percent-reduction} shows the overall percent reduction in \dll at a maximum context resulting from our experimental manipulations.
We find that obscuring coreference relations can reduce the predictive power of the LM by $16.5\%$ (Provo FF) to $42.4\%$ (OneStop GD).

\section{Discussion}

\subsection{The Dual Mechanism of Context Length}
The U-shaped predictive curve observed in Experiment 1 reveals the dual mechanism of context length for modeling human reading behavior.
Our results indicate that human reading times may reflect two distinct processing constraints simultaneously: a \emph{short-term} prediction mechanism that is influenced by working memory limits and a \emph{long-term} prediction mechanism that is required to track long-range discourse structures.
Recent computational psycholinguistic literature has predominantly focused on the former, demonstrating how severe context restrictions improve reading-time predictions~\citep{hahn-etal-2022-resource,kuribayashi-etal-2022,timkey-linzen-2023,de-varda-marelli-2024,clark-etal-2025,xu-etal-2026}.
The present study broadens this perspective by showing the latter's contribution: reading times for long documents are better predicted by surprisal from language models with longer context windows.

This dual-mechanism framework raises a fundamental question for future computational modeling: how is global discourse information compressed and represented within human working memory?
Prior research has shown memory compression for prediction at the level of individual tokens~\citep{hahn-etal-2022-resource} and syntactic structures~\citep{kajikawa-isono-ur,isono-kajikawa-2026}.
Extending these focuses to higher-level semantic representations, such as coreference chains and entity states, remains a critical open challenge.
Investigating the compression and/or representation through the lens of language models offers a promising avenue for testing cognitive theories of working memory~\citep{ericsson-kintsch-1995,kintsch-1998} that posit skilled mechanisms for processing extended narrative states.

It is noteworthy that the long-context effect was prominent in extended texts (e.g., Natural Stories, averaging 48.5 sentences per document and 21.1 words per sentence) but minimal in Provo, a shorter dataset (2.5 sentences and 13.3 words per sentence).
This suggests that reading strategies may vary by document length.
To extend this investigation across languages, multilingual reading-time datasets like MECO~\citep{siegelman-etal-2022,siegelman-etal-2025} can serve as a useful resource.
However, as they primarily consist of short texts (e.g., an average of 8.3 sentences in English MECO), developing datasets with longer narratives in multiple languages remains critical to further explore these long-range effects.

\subsection{Quantifying Discourse Contributions}
Our findings reveal that the improvement of predictive power of extended context for reading time relies heavily on structured discourse relations.
While psycholinguistic literature has long established that human readers use entity coreference to construct coherent representations during comprehension~\citep{kintsch-vandijk-1978,grosz-etal-1995,jaffe-etal-2018}, our study isolates and quantifies this process through computational modeling with LMs and surprisal theory as a linking method.
By applying a pronominalization ablation strategy, we demonstrated that obscuring structural linkages during inference degrades the predictive power of larger context windows by approximately 20\% to 40\%.
This substantial drop along with the relatively smaller decreases in our control conditions indicates that LMs do not merely treat extended context as an unstructured bag of words.
Rather, they implicitly maintain and update entity states, mirroring human discourse comprehension.

Beyond coreference, our inference-time masking methodology offers a generalizable framework to decompose other aspects of contextual influence.
By selectively disrupting specific structural linkages, researchers can apply this targeted ablation approach to quantify other factors that have been hypothesized to drive discourse representations, such as temporal event structures~\citep{zwaan-etal-1995,pustejovsky-etal-2003} or rhetorical relationships~\citep{mann-thompson-1988,zeldes-etal-2025}.
Future work should also investigate whether similar methods, applied at training time, yield similar results.
We believe that methods like ours can help move away from treating context as a monolith within computational modeling frameworks.
They have the potential to help identify and empirically validate the specific linguistic signals that drive behavioral alignment.

\section{Conclusion}
We demonstrated a U-shaped relationship between language model context size and psychometric predictive power across four English reading-time datasets.
While restricted contexts successfully capture local working memory constraints, expanded windows (500--1,000 tokens) yield the best overall fit.
Our inference-time pronominalization experiment reveals that this long-context advantage is significantly driven by entity chains, as obscuring cross-sentential coreference degrades predictive power by up to 42.4\%.
These results confirm that humans leverage global discourse relations during comprehension, demonstrating that accurate cognitive models must simultaneously capture both local working memory limits and long-range discourse integration.

\section*{Limitations}
First, our empirical findings are based exclusively on English reading-time datasets.
English relies heavily on explicit pronouns and rigid word order to establish coreference.
In pro-drop languages or morphologically rich languages, coreference relations are often mediated by verbal morphology and/or zero anaphora.
The precise nature of the U-shaped context effect and the relative contribution of entity chains to surprisal fit may vary cross-linguistically.
Future research should extend this counterfactual methodology to typologically diverse languages to verify the consistency of these dual mechanisms.

Second, while our intervention finds that LMs leverage coreference signals to improve predictability, it does not confirm that LMs and humans construct the same kind of structured representations.

Third, our counterfactual language modeling paradigm obscures coreference relations solely at inference time.
As we manipulate the input texts, the resulting evaluation data becomes slightly out-of-distribution relative to the models' naturalistic training corpora.
The models are therefore evaluated on coreference patterns they did not naturally observe during pre-training.
Future work could extend our paradigm to counterfactual training settings, where models are trained from scratch on corpora with systematically altered coreference structures to better isolate these processing effects.

Finally, our experimental pipeline relies on silver-standard annotations to identify and manipulate coreference relations.
Errors in the underlying automated coreference resolution could occasionally lead to imperfect counterfactual replacements.
Future studies could validate these findings on gold-standard human-annotated corpora to ensure that errors in automated annotation do not systematically bias the evaluation of context effects.

\section*{Ethical Considerations}
This study relies on the computational analysis of publicly available, anonymized human reading-time datasets and open-source language models. No new data from human participants was collected for this work. To the best of our knowledge, all utilized artifacts were originally collected and released in accordance with standard ethical guidelines.

\section*{AI Writing/Coding Assistance Policy}
We used generative AI tools solely for the purpose of adjusting the grammar and phrasing of the manuscript, and a coding assistant for formatting tables and figures.

\section*{Acknowledgments}
We are grateful to Ryo Yoshida, Rei Emura, Shinnosuke Isono, Sean Trott, the PICoL members at Georgetown, the members of Yuki Hirose's research group at UTokyo, and three anonymous reviewers in the ARR May 2026 cycle for their valuable comments on earlier versions of this work.
This work was supported by JSPS KAKENHI Grant Numbers 23K16938 and 26H02511.

\bibliography{custom}

@article{kajikawa-isono-ur,
  title     = {The Dual Nature of Syntactic {N}ode {C}ount: Facilitating and Inhibiting Sentence Comprehension},
  volume    = {},
  number    = {},
  journal   = {PsyArXiv preprint},
  author = {Kajikawa, Kohei and Isono, Shinnosuke},
  year      = {2026},
  publisher = {},
  pages     = {},
  url       = {https://osf.io/preprints/psyarxiv/9msby_v2}
}

@article{forster2009maze,
  title={The maze task: {M}easuring forced incremental sentence processing time},
  author={Forster, Kenneth I. and Guerrera, Christine and Elliot, Lisa},
  journal={Behavior Research Methods},
  volume={41},
  number={1},
  pages={163--171},
  year={2009},
  publisher={Springer},
  url={https://doi.org/10.3758/BRM.41.1.163}
}

@inproceedings{hale-2001,
  title     = {A Probabilistic {E}arley Parser as a Psycholinguistic Model},
  author    = {Hale, John},
  booktitle = {Second Meeting of the North {A}merican Chapter of the Association for Computational Linguistics},
  year      = {2001},
  url       = {https://aclanthology.org/N01-1021}
}

@article{smith-levy-2013,
  title    = {The effect of word predictability on reading time is logarithmic},
  journal  = {Cognition},
  volume   = {128},
  number   = {3},
  pages    = {302--319},
  year     = {2013},
  issn     = {0010--0277},
  doi      = {https://doi.org/10.1016/j.cognition.2013.02.013},
  url      = {https://www.sciencedirect.com/science/article/pii/S0010027713000413},
  author   = {Smith, Nathaniel J. and Levy, Roger}
}

@article{levy-2008,
  title    = {Expectation-based syntactic comprehension},
  journal  = {Cognition},
  volume   = {106},
  number   = {3},
  pages    = {1126--1177},
  year     = {2008},
  issn     = {0010--0277},
  doi      = {https://doi.org/10.1016/j.cognition.2007.05.006},
  url      = {https://www.sciencedirect.com/science/article/pii/S0010027707001436},
  author   = {Levy, Roger}
}

@article{frank-bod-2011,
  title     = {Insensitivity of the human sentence-processing system to hierarchical structure},
  author    = {Frank, Stefan L. and Bod, Rens},
  journal   = {Psychological Science},
  volume    = {22},
  number    = {6},
  pages     = {829--834},
  year      = {2011},
  url = {https://doi.org/10.1177/0956797611409589},
  publisher = {Sage Publications Sage CA: Los Angeles, CA}
}

@inproceedings{kennedy-etal-2003,
  title     = {{The Dundee corpus}},
  author    = {Kennedy, Alan and Hill, Robin and Pynte, Jo{\"e}l},
  booktitle = {Proceedings of the 12th European conference on eye movement},
  year      = {2003}
}

@inproceedings{radford-etal-2019,
  title  = {Language Models are Unsupervised Multitask Learners},
  author = {Radford, Alec and Wu, Jeff and Child, Rewon and Luan, David and Amodei, Dario and Sutskever, Ilya},
  year   = {2019},
  url    = {https://api.semanticscholar.org/CorpusID:160025533}
}

@manual{r-2025,
    title = {R: A Language and Environment for Statistical Computing},
    author = {{R Core Team}},
    organization = {R Foundation for Statistical Computing},
    address = {Vienna, Austria},
    year = {2025},
    url = {https://www.R-project.org/},
}

@article{shain-etal-2024,
  author   = {Shain, Cory and Meister, Clara and Pimentel, Tiago and Cotterell, Ryan and Levy, Roger},
  title    = {Large-scale evidence for logarithmic effects of word predictability on reading time},
  journal  = {Proceedings of the National Academy of Sciences},
  volume   = {121},
  number   = {10},
  pages    = {e2307876121},
  year     = {2024},
  doi      = {10.1073/pnas.2307876121},
  url      = {https://www.pnas.org/doi/abs/10.1073/pnas.2307876121},
  eprint   = {https://www.pnas.org/doi/pdf/10.1073/pnas.2307876121}
}

@article{oh-schuler-2023,
  title     = {Why Does Surprisal From Larger {T}ransformer-Based Language Models Provide a Poorer Fit to Human Reading Times?},
  author    = {Oh, Byung-Doh and Schuler, William},
  journal   = {Transactions of the Association for Computational Linguistics},
  volume    = {11},
  year      = {2023},
  address   = {Cambridge, MA},
  publisher = {MIT Press},
  url       = {https://aclanthology.org/2023.tacl-1.20},
  doi       = {10.1162/tacl_a_00548},
  pages     = {336--350}
}

@article{hahn-etal-2022-resource,
  author   = {Hahn, Michael and Futrell, Richard and Levy, Roger and Gibson, Edward},
  title    = {A resource-rational model of human processing of recursive linguistic structure},
  journal  = {Proceedings of the National Academy of Sciences},
  volume   = {119},
  number   = {43},
  pages    = {e2122602119},
  year     = {2022},
  doi      = {10.1073/pnas.2122602119},
  url      = {https://www.pnas.org/doi/abs/10.1073/pnas.2122602119},
  eprint   = {https://www.pnas.org/doi/pdf/10.1073/pnas.2122602119}
}

@article{wilcox-etal-2023-testing,
  author   = {Wilcox, Ethan G. and Pimentel, Tiago and Meister, Clara and Cotterell, Ryan and Levy, Roger P.},
  title    = {Testing the Predictions of Surprisal Theory in 11 Languages},
  journal  = {Transactions of the Association for Computational Linguistics},
  volume   = {11},
  pages    = {1451--1470},
  year     = {2023},
  month    = {12},
  issn     = {2307-387X},
  doi      = {10.1162/tacl_a_00612},
  url      = {https://doi.org/10.1162/tacl_a_00612},
  eprint   = {https://direct.mit.edu/tacl/article-pdf/doi/10.1162/tacl_a_00612/2196877/tacl_a_00612.pdf}
}

@article{futrell-etal-2021,
  author  = {Futrell, Richard and Gibson, Edward and Tily, Harry J. and Blank, Idan and Vishnevetsky, Anastasia and Piantadosi, Steven T. and Fedorenko, Evelina},
  journal = {Language Resources and Evaluation},
  title   = {The {N}atural {S}tories corpus: a reading-time corpus of {E}nglish texts containing rare syntactic constructions},
  volume  = {55},
  pages   = {63--77},
  year    = {2021},
  doi     = {10.1007/s10579-020-09503-7},
  url     = {https://doi.org/10.1007/s10579-020-09503-7}
}

@article{de-marneffe-etal-2021,
  title     = {{U}niversal {D}ependencies},
  author    = {de Marneffe, Marie-Catherine  and
               Manning, Christopher D.  and
               Nivre, Joakim  and
               Zeman, Daniel},
  journal   = {Computational Linguistics},
  volume    = {47},
  number    = {2},
  month     = jun,
  year      = {2021},
  address   = {Cambridge, MA},
  publisher = {MIT Press},
  url       = {https://aclanthology.org/2021.cl-2.11/},
  doi       = {10.1162/coli_a_00402},
  pages     = {255--308}
}

@inproceedings{wilcox-etal-2020,
  title     = {On the Predictive Power of Neural Language Models for Human Real-Time Comprehension Behavior},
  author    = {Wilcox, Ethan G. and Gauthier, Jon and Hu, Jennifer and Qian, Peng and Levy, Roger P.},
  booktitle = {Proceedings of the Annual Meeting of the Cognitive Science Society},
  year      = {2020},
  address   = {Online},
  volume    = {42},
  pages     = {1707--1713},
  url       = {https://escholarship.org/uc/item/738338tm}
}

@inproceedings{goodkind-bicknell-2018,
    title = "Predictive power of word surprisal for reading times is a linear function of language model quality",
    author = "Goodkind, Adam  and
      Bicknell, Klinton",
    editor = "Sayeed, Asad  and
      Jacobs, Cassandra  and
      Linzen, Tal  and
      van Schijndel, Marten",
    booktitle = "Proceedings of the 8th Workshop on Cognitive Modeling and Computational Linguistics ({CMCL} 2018)",
    month = jan,
    year = "2018",
    address = "Salt Lake City, Utah",
    publisher = "Association for Computational Linguistics",
    url = "https://aclanthology.org/W18-0102/",
    doi = "10.18653/v1/W18-0102",
    pages = "10--18"
}

@inproceedings{kuribayashi-etal-2022,
    title = "Context Limitations Make Neural Language Models More Human-Like",
    author = "Kuribayashi, Tatsuki  and
      Oseki, Yohei  and
      Brassard, Ana  and
      Inui, Kentaro",
    editor = "Goldberg, Yoav  and
      Kozareva, Zornitsa  and
      Zhang, Yue",
    booktitle = "Proceedings of the 2022 Conference on Empirical Methods in Natural Language Processing",
    month = dec,
    year = "2022",
    address = "Abu Dhabi, United Arab Emirates",
    publisher = "Association for Computational Linguistics",
    url = "https://aclanthology.org/2022.emnlp-main.712/",
    doi = "10.18653/v1/2022.emnlp-main.712",
    pages = "10421--10436"
}

@article{oh-etal-2022,
  author={Oh, Byung-Doh and Clark, Christian  and Schuler, William},
  title={Comparison of Structural Parsers and Neural Language Models as Surprisal Estimators},
  journal={Frontiers in Artificial Intelligence},
  volume={5},
  year={2022},
  url={https://www.frontiersin.org/journals/artificial-intelligence/articles/10.3389/frai.2022.777963},
  doi={10.3389/frai.2022.777963},
  issn={2624-8212}
}

@article{berzak-etal-2025,
  title={{OneStop}: A 360-Participant {E}nglish Eye Tracking Dataset with Different Reading Regimes},
  author={Berzak, Yevgeni and Malmaud, Jonathan and Shubi, Omer and Meiri, Yoav and Lion, Ella and Levy, Roger},
  journal={Scientific Data},
  year={2025},
  url = {https://doi.org/10.1038/s41597-025-06272-2},
  doi = {10.1038/s41597-025-06272-2}
}

@inproceedings{qi-etal-2020,
    title = "{S}tanza: A {P}ython Natural Language Processing Toolkit for Many Human Languages",
    author = "Qi, Peng  and
      Zhang, Yuhao  and
      Zhang, Yuhui  and
      Bolton, Jason  and
      Manning, Christopher D.",
    editor = "Celikyilmaz, Asli  and
      Wen, Tsung-Hsien",
    booktitle = "Proceedings of the 58th Annual Meeting of the Association for Computational Linguistics: System Demonstrations",
    month = jul,
    year = "2020",
    address = "Online",
    publisher = "Association for Computational Linguistics",
    url = "https://aclanthology.org/2020.acl-demos.14/",
    doi = "10.18653/v1/2020.acl-demos.14",
    pages = "101--108"
}

@inproceedings{oh-schuler-2024,
    title = "Leading Whitespaces of Language Models' Subword Vocabulary Pose a Confound for Calculating Word Probabilities",
    author = "Oh, Byung-Doh  and
      Schuler, William",
    editor = "Al-Onaizan, Yaser  and
      Bansal, Mohit  and
      Chen, Yun-Nung",
    booktitle = "Proceedings of the 2024 Conference on Empirical Methods in Natural Language Processing",
    month = nov,
    year = "2024",
    address = "Miami, Florida, USA",
    publisher = "Association for Computational Linguistics",
    url = "https://aclanthology.org/2024.emnlp-main.202/",
    doi = "10.18653/v1/2024.emnlp-main.202",
    pages = "3464--3472"
}

@article{gao-etal-2020,
  title={The {P}ile: An 800{GB} dataset of diverse text for language modeling},
  author={Gao, Leo and Biderman, Stella and Black, Sid and Golding, Laurence and Hoppe, Travis and Foster, Charles and Phang, Jason and He, Horace and Thite, Anish and Nabeshima, Noa and Presser, Shawn and Leahy, Connor},
  url = {https://arxiv.org/abs/2101.00027},
  journal={arXiv preprint arXiv:2101.00027},
  year={2020}
}

@article{boyce-levy-2023,
	title = {A-maze of {Natural} {Stories}: {Comprehension} and surprisal in the {Maze} task},
	volume = {2},
	issn = {2767-0279},
	shorttitle = {A-maze of {Natural} {Stories}},
	url = {https://escholarship.org/uc/item/6vh9d8zm},
	doi = {10.5070/G6011190},
	number = {1},
	urldate = {2025-10-22},
	journal = {Glossa Psycholinguistics},
	author = {Boyce, Veronica and Levy, Roger},
	month = apr,
	year = {2023}
}

@article{brms,
    title = {Bayesian Item Response Modeling in {R} with {brms} and
      {Stan}},
    author = {Paul-Christian Bürkner},
    journal = {Journal of Statistical Software},
    year = {2021},
    volume = {100},
    number = {5},
    pages = {1--54},
    doi = {10.18637/jss.v100.i05},
    encoding = {UTF-8},
  }

@inproceedings{isono-kajikawa-2026,
    title = "Syntactically-guided Information Maintenance in Sentence Comprehension",
    author = "Isono, Shinnosuke  and
      Kajikawa, Kohei",
    editor = "Bonial, Claire  and
      Berzak, Yevgeni",
    booktitle = "Proceedings of the 30th Conference on Computational Natural Language Learning",
    month = jul,
    year = "2026",
    address = "San Diego, California, USA",
    publisher = "Association for Computational Linguistics",
    url = "https://aclanthology.org/2026.conll-main.5/",
    doi = "10.18653/v1/2026.conll-main.5",
    pages = "57--69",
    ISBN = "979-8-89176-410-1"
}

@inproceedings{wolf-etal-2020-transformers,
    title = "Transformers: State-of-the-Art Natural Language Processing",
    author = "Wolf, Thomas  and
      Debut, Lysandre  and
      Sanh, Victor  and
      Chaumond, Julien  and
      Delangue, Clement  and
      Moi, Anthony  and
      Cistac, Pierric  and
      Rault, Tim  and
      Louf, Remi  and
      Funtowicz, Morgan  and
      Davison, Joe  and
      Shleifer, Sam  and
      von Platen, Patrick  and
      Ma, Clara  and
      Jernite, Yacine  and
      Plu, Julien  and
      Xu, Canwen  and
      Le Scao, Teven  and
      Gugger, Sylvain  and
      Drame, Mariama  and
      Lhoest, Quentin  and
      Rush, Alexander",
    editor = "Liu, Qun  and
      Schlangen, David",
    booktitle = "Proceedings of the 2020 Conference on Empirical Methods in Natural Language Processing: System Demonstrations",
    month = oct,
    year = "2020",
    address = "Online",
    publisher = "Association for Computational Linguistics",
    url = "https://aclanthology.org/2020.emnlp-demos.6/",
    doi = "10.18653/v1/2020.emnlp-demos.6",
    pages = "38--45"
}

@article{luke-christianson-2018,
    author = {Luke, Steven G. and Christianson, Kiel},
    title = {The Provo Corpus: A large eye-tracking corpus with predictability norms},
    journal = {Behavior Research Methods},
    year = {2018},
    pages = {826--833},
    volume = {50},
    number = {2},
    url = {https://doi.org/10.3758/s13428-017-0908-4}    
}

@inproceedings{timkey-linzen-2023,
    title = "A Language Model with Limited Memory Capacity Captures Interference in Human Sentence Processing",
    author = "Timkey, William  and
      Linzen, Tal",
    editor = "Bouamor, Houda  and
      Pino, Juan  and
      Bali, Kalika",
    booktitle = "Findings of the Association for Computational Linguistics: EMNLP 2023",
    month = dec,
    year = "2023",
    address = "Singapore",
    publisher = "Association for Computational Linguistics",
    url = "https://aclanthology.org/2023.findings-emnlp.582/",
    doi = "10.18653/v1/2023.findings-emnlp.582",
    pages = "8705--8720"
}

@inproceedings{de-varda-marelli-2024,
    title = "Locally Biased Transformers Better Align with Human Reading Times",
    author = "De Varda, Andrea  and
      Marelli, Marco",
    editor = "Kuribayashi, Tatsuki  and
      Rambelli, Giulia  and
      Takmaz, Ece  and
      Wicke, Philipp  and
      Oseki, Yohei",
    booktitle = "Proceedings of the Workshop on Cognitive Modeling and Computational Linguistics",
    month = aug,
    year = "2024",
    address = "Bangkok, Thailand",
    publisher = "Association for Computational Linguistics",
    url = "https://aclanthology.org/2024.cmcl-1.3/",
    doi = "10.18653/v1/2024.cmcl-1.3",
    pages = "30--36"
}

@inproceedings{clark-etal-2025,
    title = "Linear Recency Bias During Training Improves Transformers' Fit to Reading Times",
    author = "Clark, Christian  and
      Oh, Byung-Doh  and
      Schuler, William",
    editor = "Rambow, Owen  and
      Wanner, Leo  and
      Apidianaki, Marianna  and
      Al-Khalifa, Hend  and
      Eugenio, Barbara Di  and
      Schockaert, Steven",
    booktitle = "Proceedings of the 31st International Conference on Computational Linguistics",
    month = jan,
    year = "2025",
    address = "Abu Dhabi, UAE",
    publisher = "Association for Computational Linguistics",
    url = "https://aclanthology.org/2025.coling-main.517/",
    pages = "7735--7747"
}

@inproceedings{jaffe-etal-2018,
    title = "Coreference and Focus in Reading Times",
    author = "Jaffe, Evan  and
      Shain, Cory  and
      Schuler, William",
    editor = "Sayeed, Asad  and
      Jacobs, Cassandra  and
      Linzen, Tal  and
      van Schijndel, Marten",
    booktitle = "Proceedings of the 8th Workshop on Cognitive Modeling and Computational Linguistics ({CMCL} 2018)",
    month = jan,
    year = "2018",
    address = "Salt Lake City, Utah",
    publisher = "Association for Computational Linguistics",
    url = "https://aclanthology.org/W18-0101/",
    doi = "10.18653/v1/W18-0101",
    pages = "1--9"
}

@inproceedings{kuribayashi-etal-2024,
    title = "Psychometric Predictive Power of Large Language Models",
    author = "Kuribayashi, Tatsuki  and
      Oseki, Yohei  and
      Baldwin, Timothy",
    editor = "Duh, Kevin  and
      Gomez, Helena  and
      Bethard, Steven",
    booktitle = "Findings of the Association for Computational Linguistics: NAACL 2024",
    month = jun,
    year = "2024",
    address = "Mexico City, Mexico",
    publisher = "Association for Computational Linguistics",
    url = "https://aclanthology.org/2024.findings-naacl.129/",
    doi = "10.18653/v1/2024.findings-naacl.129",
    pages = "1983--2005"
}

@article{kuribayashi-etal-2025,
    title = "Large Language Models Are Human-Like Internally",
    author = "Kuribayashi, Tatsuki  and
      Oseki, Yohei  and
      Taieb, Souhaib Ben  and
      Inui, Kentaro  and
      Baldwin, Timothy",
    journal = "Transactions of the Association for Computational Linguistics",
    volume = "13",
    year = "2025",
    address = "Cambridge, MA",
    publisher = "MIT Press",
    url = "https://aclanthology.org/2025.tacl-1.78/",
    doi = "10.1162/tacl.a.58",
    pages = "1743--1766"
}

@inproceedings{tsipidi-etal-2025-harmonic,
    title = "The Harmonic Structure of Information Contours",
    author = "Tsipidi, Eleftheria  and
      Kiegeland, Samuel  and
      Nowak, Franz  and
      Xu, Tianyang  and
      Wilcox, Ethan  and
      Warstadt, Alex  and
      Cotterell, Ryan  and
      Giulianelli, Mario",
    editor = "Che, Wanxiang  and
      Nabende, Joyce  and
      Shutova, Ekaterina  and
      Pilehvar, Mohammad Taher",
    booktitle = "Proceedings of the 63rd Annual Meeting of the Association for Computational Linguistics (Volume 1: Long Papers)",
    month = jul,
    year = "2025",
    address = "Vienna, Austria",
    publisher = "Association for Computational Linguistics",
    url = "https://aclanthology.org/2025.acl-long.1527/",
    doi = "10.18653/v1/2025.acl-long.1527",
    pages = "31636--31659",
    ISBN = "979-8-89176-251-0"
}

@article{oh-linzen-2026,
  title={To model human linguistic prediction, make {LLM}s less superhuman},
  author={Oh, Byung-Doh and Linzen, Tal},
  journal={Trends in Cognitive Sciences},
  url = {https://doi.org/10.1016/j.tics.2026.05.008},
  year={2026},
  publisher={Elsevier}
}

@inproceedings{tsipidi-etal-2024,
    title = "Surprise! {U}niform {I}nformation {D}ensity Isn{'}t the Whole Story: Predicting Surprisal Contours in Long-form Discourse",
    author = "Tsipidi, Eleftheria  and
      Nowak, Franz  and
      Cotterell, Ryan  and
      Wilcox, Ethan  and
      Giulianelli, Mario  and
      Warstadt, Alex",
    editor = "Al-Onaizan, Yaser  and
      Bansal, Mohit  and
      Chen, Yun-Nung",
    booktitle = "Proceedings of the 2024 Conference on Empirical Methods in Natural Language Processing",
    month = nov,
    year = "2024",
    address = "Miami, Florida, USA",
    publisher = "Association for Computational Linguistics",
    url = "https://aclanthology.org/2024.emnlp-main.1047/",
    doi = "10.18653/v1/2024.emnlp-main.1047",
    pages = "18820--18836"
}

@inproceedings{liu-etal-2024-mscaw,
    title = "{MSCAW}-coref: Multilingual, Singleton and Conjunction-Aware Word-Level Coreference Resolution",
    author = "Liu, Houjun  and
      Bauer, John  and
      D{'}Oosterlinck, Karel  and
      Potts, Christopher  and
      Manning, Christopher D.",
    editor = "Ogrodniczuk, Maciej  and
      Nedoluzhko, Anna  and
      Poesio, Massimo  and
      Pradhan, Sameer  and
      Ng, Vincent",
    booktitle = "Proceedings of the Seventh Workshop on Computational Models of Reference, Anaphora and Coreference",
    month = nov,
    year = "2024",
    address = "Miami",
    publisher = "Association for Computational Linguistics",
    url = "https://aclanthology.org/2024.crac-1.4/",
    doi = "10.18653/v1/2024.crac-1.4",
    pages = "33--40"
}

@article{goldstein-etal-2022,
    author = {Goldstein, Ariel and Zada, Zaid and Buchnik, Eliav and Schain, Mariano and Price, Amy and Aubrey, Bobbi and Nastase, Samuel A. and Feder, Amir and Emanuel, Dotan and Cohen, Alon and Jansen, Aren and Gazula, Harshvardhan and Choe, Gina and Rao, Aditi and Kim, Catherine and Casto, Colton and Fanda, Lora and Doyle, Werner and Friedman, Daniel and Dugan, Patricia and Melloni, Lucia and Reichart, Roi and Devore, Sasha and Flinker, Adeen and Hasenfratz, Liat and Levy, Omer and Hassidim, Avinatan and Brenner, Michael and Matias, Yossi and Norman, Kenneth A. and Devinsky, Orrin and Hasson, Uri},
    title = {Shared computational principles for language processing in humans and deep language models},
    journal = {Nature Neuroscience},
    year = {2022},
    pages ={369--380},
    volume = {25},
    url = {https://doi.org/10.1038/s41593-022-01026-4}
}

@inproceedings{brown-etal-2020,
 author = {Brown, Tom and Mann, Benjamin and Ryder, Nick and Subbiah, Melanie and Kaplan, Jared D and Dhariwal, Prafulla and Neelakantan, Arvind and Shyam, Pranav and Sastry, Girish and Askell, Amanda and Agarwal, Sandhini and Herbert-Voss, Ariel and Krueger, Gretchen and Henighan, Tom and Child, Rewon and Ramesh, Aditya and Ziegler, Daniel and Wu, Jeffrey and Winter, Clemens and Hesse, Chris and Chen, Mark and Sigler, Eric and Litwin, Mateusz and Gray, Scott and Chess, Benjamin and Clark, Jack and Berner, Christopher and McCandlish, Sam and Radford, Alec and Sutskever, Ilya and Amodei, Dario},
 booktitle = {Advances in Neural Information Processing Systems},
 editor = {H. Larochelle and M. Ranzato and R. Hadsell and M.F. Balcan and H. Lin},
 pages = {1877--1901},
 publisher = {Curran Associates, Inc.},
 title = {Language Models are Few-Shot Learners},
 url = {https://proceedings.neurips.cc/paper_files/paper/2020/file/1457c0d6bfcb4967418bfb8ac142f64a-Paper.pdf},
 volume = {33},
 year = {2020}
}

@article{frank-goodman-2026,
   author = "Frank, Michael C. and Goodman, Noah D.",
   title = "Cognitive Modeling Using Artificial Intelligence", 
   journal= "Annual Review of Psychology",
   year = "2026",
   volume = "77",
   pages = "543-566",
   doi = "https://doi.org/10.1146/annurev-psych-030625-040748",
   url = "https://www.annualreviews.org/content/journals/10.1146/annurev-psych-030625-040748",
   publisher = "Annual Reviews",
   issn = "1545-2085",
  }

@article{grosz-etal-1995,
    title = "{C}entering: A Framework for Modeling the Local Coherence of Discourse",
    author = "Grosz, Barbara J.  and
      Joshi, Aravind K.  and
      Weinstein, Scott",
    editor = "Hirschberg, Julia",
    journal = "Computational Linguistics",
    volume = "21",
    number = "2",
    year = "1995",
    address = "Cambridge, MA",
    publisher = "MIT Press",
    url = "https://aclanthology.org/J95-2003/",
    pages = "203--225"
}

@article{kintsch-vandijk-1978,
    author = {Kintsch, Walter and van Dijk, Teun A.},
    year = {1978},
    title = {Toward a model of text comprehension and production},
    journal = {Psychological Review},
    volume = {85},
    number = {5},
    pages = {363--394},
    url = {https://doi.org/10.1037/0033-295X.85.5.363}
}

@article{mann-thompson-1988,
    title = {{Rhetorical Structure Theory}: Toward a functional theory of text organization},
    author = {Mann, William C. and Thompson, Sandra A.},
    pages = {243--281},
    volume = {8},
    number = {3},
    journal = {Text - Interdisciplinary Journal for the Study of Discourse},
    doi = {doi:10.1515/text.1.1988.8.3.243},
    year = {1988}
}

@article{zeldes-etal-2025,
    title = "e{RST}: A Signaled Graph Theory of Discourse Relations and Organization",
    author = "Zeldes, Amir  and
      Aoyama, Tatsuya  and
      Liu, Yang Janet  and
      Peng, Siyao  and
      Das, Debopam  and
      Gessler, Luke",
    journal = "Computational Linguistics",
    volume = "51",
    number = "1",
    month = mar,
    year = "2025",
    address = "Cambridge, MA",
    publisher = "MIT Press",
    url = "https://aclanthology.org/2025.cl-1.3/",
    doi = "10.1162/coli_a_00538",
    pages = "23--72"
}

@incollection{pustejovsky-etal-2003,
  title={Time{ML}: Robust specification of event and temporal expressions in text.},
  author={Pustejovsky, James and Castano, Jos{\'e} and Ingria, Robert and Sauri, Roser and Gaizauskas, Robert and Setzer, Andrea and Katz, Graham and Radev, Dragomir},
  booktitle ={Proceedings of New Directions in Question Answering},
  year={2003},
  url = {https://aaai.org/papers/0005-ss03-07-005-timeml-robust-specification-of-event-and-temporal-expressions-in-text/}
}

@article{zwaan-etal-1995,
    author = {Zwaan, Rolf A. and Magliano, Joseph P. and Graesser, Arthur C.},
    title = {Dimensions of situation model construction in narrative comprehension},
    journal = {Journal of Experimental Psychology: Learning, Memory, and Cognition},
    year = {1995},
    volume = {21},
    number = {2},
    pages = {386--397},
    url = {https://doi.org/10.1037/0278-7393.21.2.386}
}

@inproceedings{clark-etal-2019-bert,
    title = "What Does {BERT} Look at? An Analysis of {BERT}{'}s Attention",
    author = "Clark, Kevin  and
      Khandelwal, Urvashi  and
      Levy, Omer  and
      Manning, Christopher D.",
    editor = "Linzen, Tal  and
      Chrupa{\l}a, Grzegorz  and
      Belinkov, Yonatan  and
      Hupkes, Dieuwke",
    booktitle = "Proceedings of the 2019 ACL Workshop BlackboxNLP: Analyzing and Interpreting Neural Networks for NLP",
    month = aug,
    year = "2019",
    address = "Florence, Italy",
    publisher = "Association for Computational Linguistics",
    url = "https://aclanthology.org/W19-4828/",
    doi = "10.18653/v1/W19-4828",
    pages = "276--286"
}

@inproceedings{sorodoc-etal-2020-probing,
    title = "Probing for Referential Information in Language Models",
    author = "Sorodoc, Ionut-Teodor  and
      Gulordava, Kristina  and
      Boleda, Gemma",
    editor = "Jurafsky, Dan  and
      Chai, Joyce  and
      Schluter, Natalie  and
      Tetreault, Joel",
    booktitle = "Proceedings of the 58th Annual Meeting of the Association for Computational Linguistics",
    month = jul,
    year = "2020",
    address = "Online",
    publisher = "Association for Computational Linguistics",
    url = "https://aclanthology.org/2020.acl-main.384/",
    doi = "10.18653/v1/2020.acl-main.384",
    pages = "4177--4189"
}

@inproceedings{tenney-etal-2019-bert,
    title = "{BERT} Rediscovers the Classical {NLP} Pipeline",
    author = "Tenney, Ian  and
      Das, Dipanjan  and
      Pavlick, Ellie",
    editor = "Korhonen, Anna  and
      Traum, David  and
      M{\`a}rquez, Llu{\'i}s",
    booktitle = "Proceedings of the 57th Annual Meeting of the Association for Computational Linguistics",
    month = jul,
    year = "2019",
    address = "Florence, Italy",
    publisher = "Association for Computational Linguistics",
    url = "https://aclanthology.org/P19-1452/",
    doi = "10.18653/v1/P19-1452",
    pages = "4593--4601"
}

@inbook{jm3,
    author = "Daniel Jurafsky and James H. Martin",
    booktitle = "Speech and Language Processing: An Introduction to Natural Language Processing, Computational Linguistics, and Recognition, with Language Models",
    year =  "2026",
    title = "Coreference Resolution and
Entity Linking",
    url = 
    "https://web.stanford.edu/~jurafsky/slp3/23.pdf",
    note = "Online manuscript released January 6, 2026",
  edition =         "3rd",
}

@inproceedings{karttunen-1969-discourse-referents,
    title = "Discourse Referents",
    author = "Karttunen, Lauri",
    booktitle = "{I}nternational {C}onference on {C}omputational {L}inguistics {COLING} 1969: Preprint No. 70",
    month = sep,
    year = "1969",
    address = {S{\r{a}}nga S{\"a}by, Sweden},
    url = "https://aclanthology.org/C69-7001/"
}

@inproceedings{nedoluzhko-etal-2022-corefud,
    title = "{C}oref{UD} 1.0: Coreference Meets {U}niversal {D}ependencies",
    author = "Nedoluzhko, Anna  and
      Nov{\'a}k, Michal  and
      Popel, Martin  and
      {\v{Z}}abokrtsk{\'y}, Zden{\v{e}}k  and
      Zeldes, Amir  and
      Zeman, Daniel",
    editor = "Calzolari, Nicoletta  and
      B{\'e}chet, Fr{\'e}d{\'e}ric  and
      Blache, Philippe  and
      Choukri, Khalid  and
      Cieri, Christopher  and
      Declerck, Thierry  and
      Goggi, Sara  and
      Isahara, Hitoshi  and
      Maegaard, Bente  and
      Mariani, Joseph  and
      Mazo, H{\'e}l{\`e}ne  and
      Odijk, Jan  and
      Piperidis, Stelios",
    booktitle = "Proceedings of the Thirteenth Language Resources and Evaluation Conference",
    month = jun,
    year = "2022",
    address = "Marseille, France",
    publisher = "European Language Resources Association",
    url = "https://aclanthology.org/2022.lrec-1.520/",
    pages = "4859--4872"
}

@article{hobbs1979coherence,
  title={Coherence and coreference},
  author={Hobbs, Jerry R.},
  journal={Cognitive Science},
  volume={3},
  number={1},
  pages={67--90},
  year={1979},
  url = {https://www.sciencedirect.com/science/article/pii/S0364021379800439},
  publisher={Wiley Online Library}
}

@misc{deepseekai2026deepseekv4,
      title={DeepSeek-V4: Towards Highly Efficient Million-Token Context Intelligence},
      author={DeepSeek-AI},
      year={2026},
      url={https://huggingface.co/deepseek-ai/DeepSeek-V4-Pro}
}

@article{ericsson-kintsch-1995,
    author = {Ericsson, K. Anders and Kintsch, Walter},
    title = {Long-term working memory},
    journal = {Psychological Review},
    year = {1995},
    pages = {211--245},
    volume = {102},
    number = {2},
    url = {https://doi.org/10.1037/0033-295X.102.2.211}
}

@book{kintsch-1998,
    author = {Kintsch, Walter},
    title = {Comprehension: A paradigm for cognition},
    publisher = {Cambridge University Press},
    year = {1998}
}

@article{daneman-carpenter-1980,
title = {Individual differences in working memory and reading},
journal = {Journal of Verbal Learning and Verbal Behavior},
year = {1980},
volume = {19},
number = {4},
pages = {450--466},
year = {1980},
issn = {0022-5371},
doi = {https://doi.org/10.1016/S0022-5371(80)90312-6},
url = {https://www.sciencedirect.com/science/article/pii/S0022537180903126},
author = {Daneman, Meredyth and Carpenter, Patricia A.},
}

@inproceedings{xu-etal-2026,
    title = "Memory efficiency and resource-rational encoding in sentence processing",
    author = "Xu, Weijie  and
      Dillon, Brian  and
      Futrell, Richard",
    editor = "Liakata, Maria  and
      Moreira, Viviane P.  and
      Zhang, Jiajun  and
      Jurgens, David",
    booktitle = "Proceedings of the 64th Annual Meeting of the {A}ssociation for {C}omputational {L}inguistics (Volume 1: Long Papers)",
    month = jul,
    year = "2026",
    address = "San Diego, California, United States",
    publisher = "Association for Computational Linguistics",
    url = "https://aclanthology.org/2026.acl-long.1550/",
    doi = "10.18653/v1/2026.acl-long.1550",
    pages = "33603--33618",
    ISBN = "979-8-89176-390-6"
}

@article{siegelman-etal-2022,
    author = {Siegelman, Noam and Schroeder, Sascha and Acartürk, Cengiz and Ahn, Hee-Don and Alexeeva, Svetlana and Amenta, Simona and Bertram, Raymond and Bonandrini, Rolando and Brysbaert, Marc and Chernova, Daria and Da Fonseca, Sara Maria and Dirix, Nicolas and Duyck, Wouter and Fella, Argyro and Frost, Ram and Gattei, Carolina A. and Kalaitzi, Areti and Kwon, Nayoung and Lõo, Kaidi and Marelli, Marco and Papadopoulos, Timothy C. and Protopapas, Athanassios and Savo, Satu and Shalom, Diego E. and Slioussar, Natalia and Stein, Roni and Sui, Longjiao and Taboh, Analí and Tønnesen, Veronica and Usal, Kerem Alp and Kuperman, Victor},
    title = {Expanding horizons of cross-linguistic research on reading: The Multilingual Eye-movement Corpus (MECO)},
    journal = {Behavior Research Methods},
    year = {2022},
    pages = {2843--2863},
    volume = {54},
    issue = {6},
    url = {https://doi.org/10.3758/s13428-021-01772-6}
}

@article{siegelman-etal-2025,
    author = {Siegelman, Noam and Schroeder, Sascha and Bao, Yaqian Borogjoon and Acartürk, Cengiz and Agrawal, Niket and Bolliger, Lena S. and Brasser, Jan and Campos-Rojas, César and Drieghe, Denis and Filipović Đurđević, Dušica and Goldina, Sofya and Ibáñez Orellana, Romualdo and Jäger, Lena A. and Jóhannesson, Ómar I. and Khare, Anurag and Kharlamov, Nik and Knudsen, Hanne B. S. and Kristjánsson, Árni and Lee, Charlotte E. and Lee, Jun Ren and Leite, Marina P. T. and Mancini, Simona and Mihajlović, Nataša and Mišić, Ksenija and Orekhova, Miloslava and Parshina, Olga and Popović Stijačić, Milica and Protopapas, Athanassios and Reich, David R. and Rimzhim, Anurag and Rothe-Neves, Rui and Sá, Thais M. M. and Santana-Covarrubias, Andrea and Sekerina, Irina and Sigurdardottir, Heida M. and Smirnova, Anna and Srivastava, Priyanka and Teixeira, Elisangela N. and Ugrinic, Ivana and Usal, Kerem Alp and Vakulya, Karolina and Verma, Ark and Vieira, João M. M. and Wu, Denise H. and Xue, Jin and Zdravković, Sunčica and Zhuo, Junjing and Ziaka, Laoura and Kuperman, Victor},
    title = {Wave 2 of the Multilingual Eye-Movement Corpus (MECO): New text reading data across languages},
    journal = {Scientific Data},
    year = {2025},
    volume = {12},
    issue = {1},
    pages = {1183},
    url = {https://doi.org/10.1038/s41597-025-05453-3}
}

@article{zhang-etal-2022,
  title={Opt: Open pre-trained {T}ransformer language models},
  author={Zhang, Susan and Roller, Stephen and Goyal, Naman and Artetxe, Mikel and Chen, Moya and Chen, Shuohui and Dewan, Christopher and Diab, Mona and Li, Xian and Lin, Xi Victoria and Mihaylov, Todor and Ott, Myle and Shleifer, Sam and Shuster, Kurt and Simig, Daniel and Koura, Punit Singh and Sridhar, Anjali and Wang, Tianlu and Zettlemoyer, Luke},
  journal={arXiv preprint arXiv:2205.01068},
  year={2022}
}

@software{gpt-neo,
  author       = {Black, Sid and
                  Gao, Leo and
                  Wang, Phil and
                  Leahy, Connor and
                  Biderman, Stella},
  title        = {{GPT-Neo: Large Scale Autoregressive Language 
                   Modeling with Mesh-Tensorflow}},
  month        = mar,
  year         = 2021,
  note         = {{If you use this software, please cite it using 
                   these metadata.}},
  publisher    = {Zenodo},
  version      = {1.0},
  doi          = {10.5281/zenodo.5297715},
  url          = {https://doi.org/10.5281/zenodo.5297715}
}

@misc{gpt-j,
  author = {Wang, Ben and Komatsuzaki, Aran},
  title = {{GPT-J-6B: A 6 Billion Parameter Autoregressive Language Model}},
  howpublished = {\url{https://github.com/kingoflolz/mesh-transformer-jax}},
  year = 2021,
  month = May
}

\appendix
\crefalias{section}{appendix}
\crefalias{subsection}{appendix}

\section{Artifacts}
\subsection{Language Models}
\Cref{tab:lm} lists the language models used in \cref{sec:exp1,sec:exp2}.

\begin{tcolorbox}[float*,width=\textwidth,colback=gray!5!white,
    colframe=gray!75!black,
    label={box:prompt_template},title=Prompt template used for batch pronominalization. Braced fields denote values populated at inference time.]
\begin{verbatim}
System:
You are a linguistic expert. You will receive sentences with placeholders like
[[TARGET_N]]. Your task is to choose exactly one appropriate English pronoun for
each placeholder. Use sentence position to choose case: subject, object,
possessive, or reflexive. Refer to the provided Context only to determine number,
animacy, and likely gender. Never return the original Context text as a value.
Return ONLY a valid JSON object. It must have this shape:
{"results": [{"sentence_index": 0, "pronoun_mapping": {"[[TARGET_0]]": "it"}}]}.
Every placeholder in Context must appear exactly once in pronoun_mapping.

User:
Sentence {sentence_index}: {masked_sentence}
Context: [[TARGET_0]]: {original_noun_phrase_0},
[[TARGET_1]]: {original_noun_phrase_1}, ...

Sentence {sentence_index}: {masked_sentence}
Context: [[TARGET_0]]: {original_noun_phrase_0},
[[TARGET_1]]: {original_noun_phrase_1}, ...
\end{verbatim}
\end{tcolorbox}
\begin{table*}[ht]
    \centering
    \setlength{\tabcolsep}{4pt}
    \begin{tabular}{lll}
        \toprule
        \textbf{Model} & \textbf{URL} & \textbf{\#Params}\\
        \midrule
        GPT2-small (\gptsmall) & \url{https://huggingface.co/openai-community/gpt2} & 124M \\
        GPT2-medium (\gptmedium) & \url{https://huggingface.co/openai-community/gpt2-medium} & 355M \\
        GPT2-large (\gptlarge) & \url{https://huggingface.co/openai-community/gpt2-large} & 774M \\
        GPT2-xl (\gptxl) & \url{https://huggingface.co/openai-community/gpt2-xl} & 1.5B \\
        GPT-Neo 125M & \url{https://huggingface.co/EleutherAI/gpt-neo-125m} & 125M \\
        GPT-Neo 1.3B & \url{https://huggingface.co/EleutherAI/gpt-neo-1.3B} & 1.3B \\
        GPT-Neo 2.7B & \url{https://huggingface.co/EleutherAI/gpt-neo-2.7B} & 2.7B \\
        GPT-J 6B & \url{https://huggingface.co/EleutherAI/gpt-j-6b} & 6B \\
        OPT 125M & \url{https://huggingface.co/facebook/opt-125m} & 125M \\
        OPT 350M & \url{https://huggingface.co/facebook/opt-350m} & 350M \\
        OPT 1.3B & \url{https://huggingface.co/facebook/opt-1.3b} & 1.3B \\
        OPT 2.7B & \url{https://huggingface.co/facebook/opt-2.7b} & 2.7B \\
        OPT 6.7B & \url{https://huggingface.co/facebook/opt-6.7b} & 6.7B \\
        OPT 13B & \url{https://huggingface.co/facebook/opt-13b} & 13B \\
        OPT 30B & \url{https://huggingface.co/facebook/opt-30b} & 30B \\
        OPT 66B & \url{https://huggingface.co/facebook/opt-66b} & 66B \\
        \bottomrule
    \end{tabular}
    \caption{Pretrained language models used in this paper.}
    \label{tab:lm}
\end{table*}

\subsection{Data and Tools}
\cref{tab:licenses} lists the datasets and tools used in \cref{sec:exp1,sec:exp2}.
We used these artifacts according to their licenses.
For computing surprisal, we used a single NVIDIA H100 GPU (80GB) for several hours.

\begin{table*}[ht]
    \centering
    \setlength{\tabcolsep}{4pt}
    \begin{tabular}{ll}
        \toprule
        \textbf{Artifact} & \textbf{License} \\
        \midrule
        Brown~\citep{smith-levy-2013} & CC BY 3.0 \\
        Natural Stories~\citep{futrell-etal-2021} & CC BY-NC-SA 4.0 \\
        OneStop~\citep{berzak-etal-2025} & CC BY 4.0 \\
        Provo~\citep{luke-christianson-2018} & CC BY 4.0 \\
        Transformers~\citep{wolf-etal-2020-transformers} & Apache 2.0 \\
        Stanza~\citep{qi-etal-2020,liu-etal-2024-mscaw} & Apache 2.0 \\
        brms~\citep{brms} & GNU General Public License v2.0 \\
        \mbox{DeepSeek-V4-Flash} & MIT License \\
        Gemini-3.6-Thinking & Proprietary \\
        \bottomrule
    \end{tabular}
    \caption{Artifacts and their licenses used in this paper.}
    \label{tab:licenses}
\end{table*}

\section{Mean Surprisal}\label{app:meansurp}

\begin{figure*}
    \centering
    \includegraphics[width=0.8\linewidth]{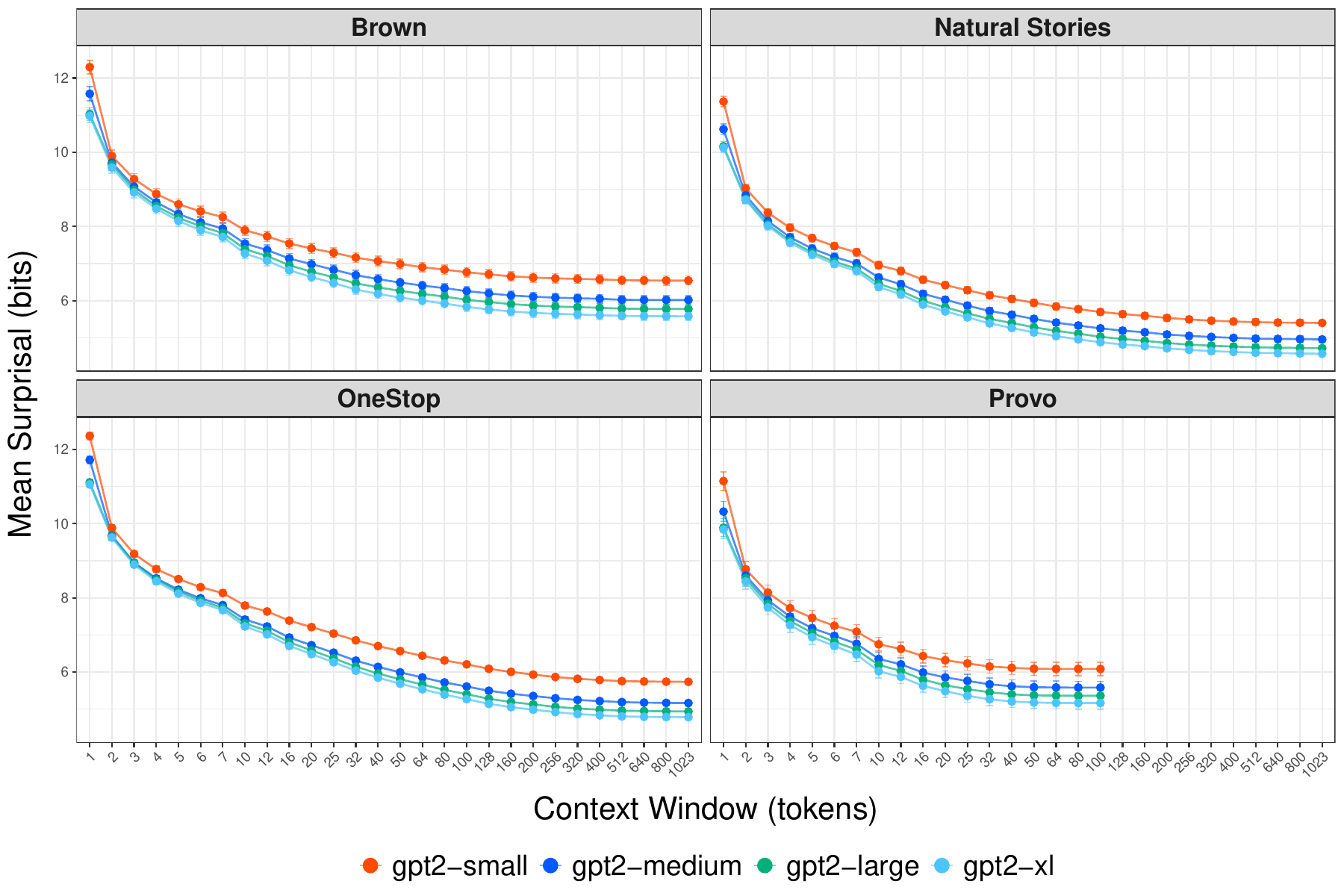}
    \caption{Mean surprisal as a function of context window size up to maximum length across four reading-time datasets. Each point represents the mean value and error bars show 95\% confidence interval.}
    \label{fig:meansurp}
\end{figure*}

\Cref{fig:meansurp} shows the mean surprisal values for each dataset across all context sizes examined in this study.

\section{Experiment 1 Results on Other Model Families}\label{app:ns_biggermodels}

In addition to the GPT-2 series in Experiment 1 (\cref{sec:exp1}), we evaluate the predictive power of other language model families, specifically the GPT-Neo~\citep{gpt-neo}, GPT-J~\citep{gpt-j}, and OPT~\citep{zhang-etal-2022} series, on Natural Stories SPR to ensure that our findings generalize across different architectures.

\begin{figure}[htbp]
    \centering
    \includegraphics[width=\linewidth]{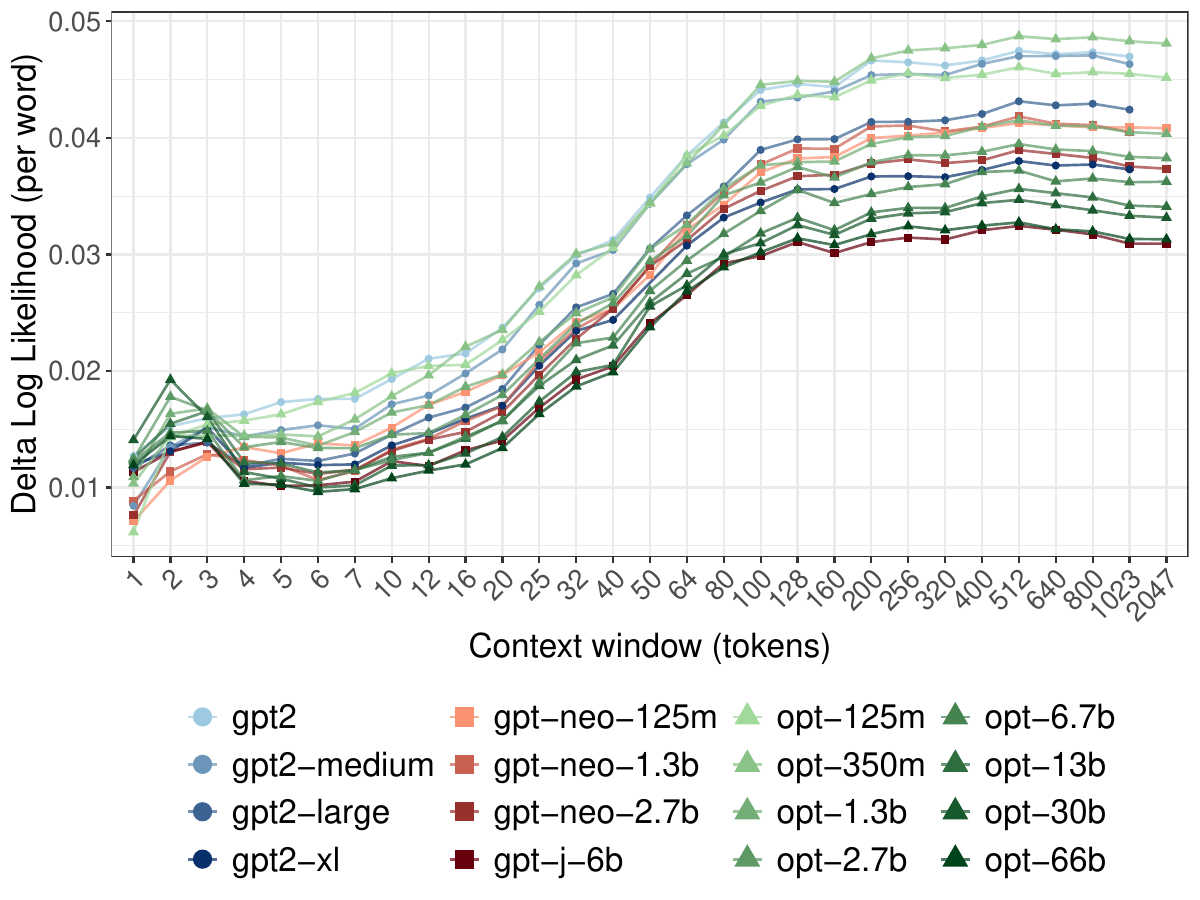}
    \caption{Predictive power (\dll) across varying context window sizes for the GPT-2, GPT-Neo, GPT-J, and OPT model series on Natural Stories SPR.}
    \label{fig:biggermodels}
\end{figure}

\Cref{fig:biggermodels} shows the result.
For models with a maximum context length of 2,048 tokens, we also evaluate and plot the \dll at their maximum context length. 
Consistent with our main findings, we observe a U-, or sometimes S-shaped pattern regardless of the model family.

\section{Full Results for Experiment 2 across All Models}\label{app:delta_breakdown}

\begin{figure*}
    \centering
    \includegraphics[width=\linewidth]{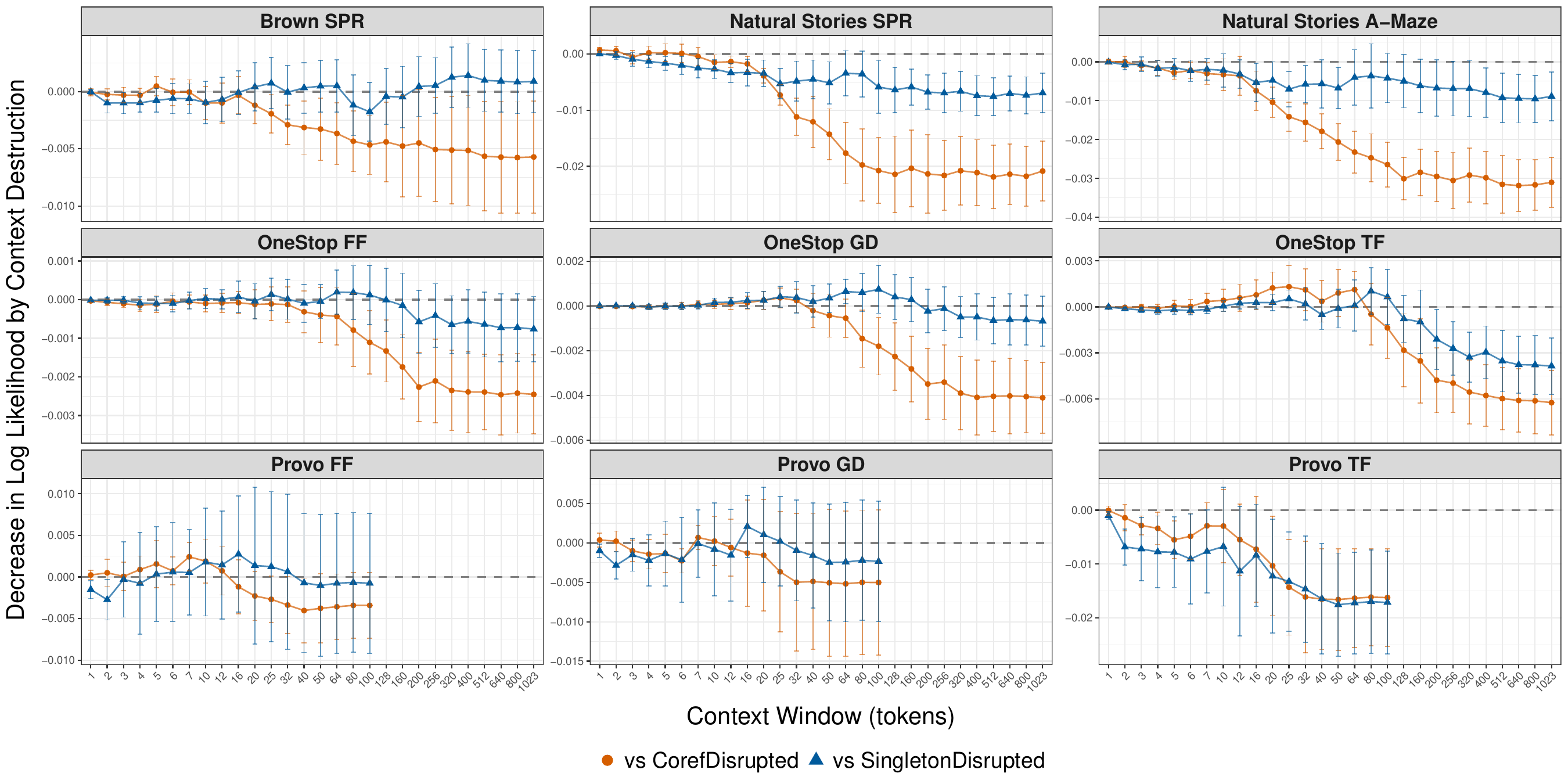}
    \caption{Difference of predictive power between the original version reported in \cref{fig:dll} and \baseline and between the original version and \linking across multiple context sizes in \gptmedium.}
    \label{fig:delta_med}
\end{figure*}

\begin{figure*}
    \centering
    \includegraphics[width=\linewidth]{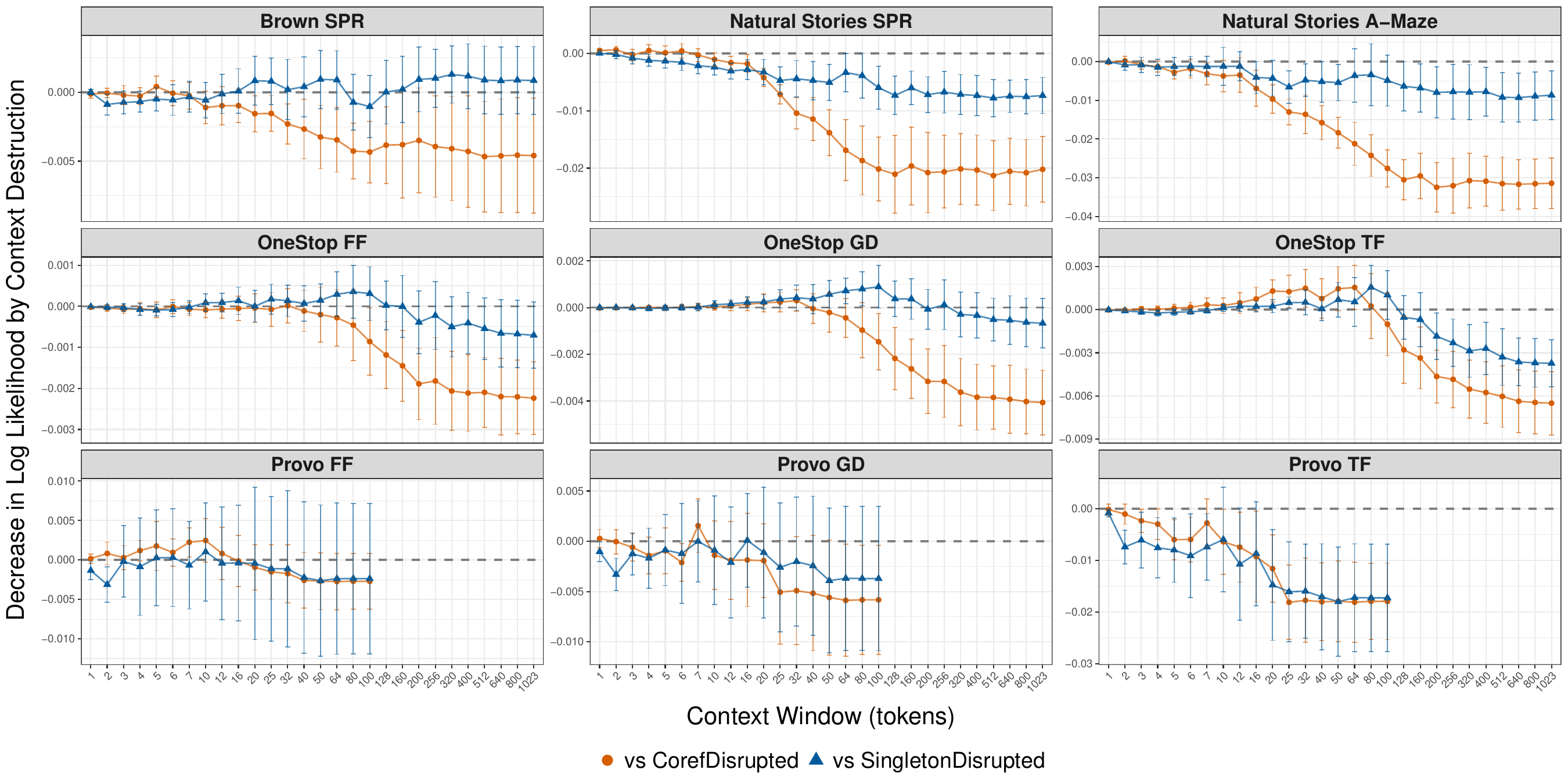}
    \caption{Difference of predictive power between the original version reported in \cref{fig:dll} and \baseline and between the original version and \linking across multiple context sizes in \gptlarge.}
    \label{fig:delta_large}
\end{figure*}

\begin{figure*}
    \centering
    \includegraphics[width=\linewidth]{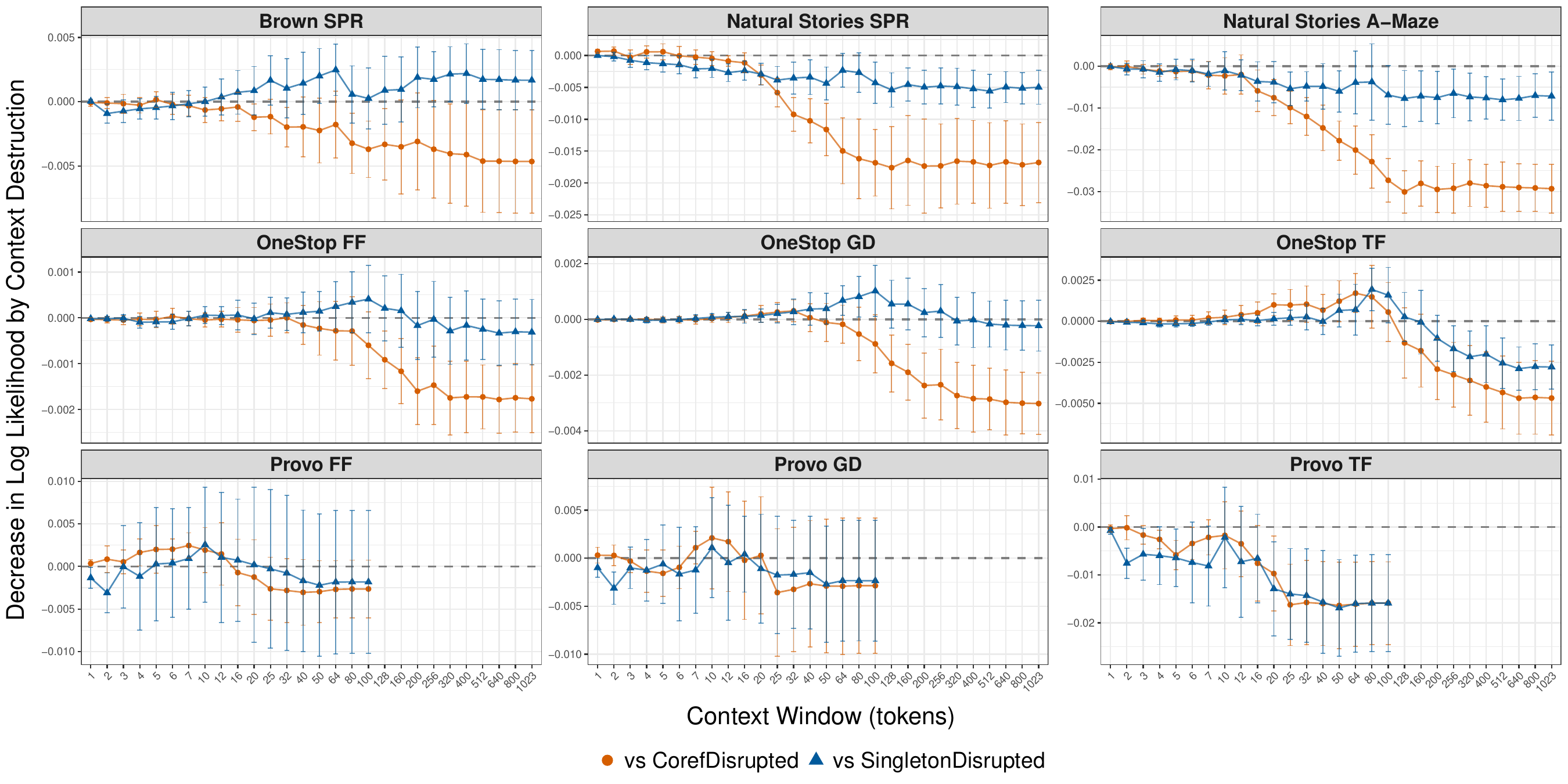}
    \caption{Difference of predictive power between the original version reported in \cref{fig:dll} and \baseline and between the original version and \linking across multiple context sizes in \gptxl.}
    \label{fig:delta_xl}
\end{figure*}

\Cref{fig:delta_med,fig:delta_large,fig:delta_xl} illustrate the results of Experiment 2 in \cref{sec:exp2} for GPT2-\gptmedium, \gptlarge, and \gptxl, respectively.
Results are consistent with those reported in the main body of the text: The \linking condition results in a substantial drop of predictive power relative to the baseline. The \baseline condition results in no or a smaller reduction in predictive power. As with our main results, Provo is an exception, likely due to its shorter context lengths.

\section{Prompt template for substituting an entity span with a pronoun}\label{app:prompt}

We used the following prompt template to obtain pronoun replacements for masked noun phrases. Each span of contiguous masked tokens is replaced with a placeholder of the form \texttt{[[TARGET\_N]]}. Each input sentence contained one or more placeholders. The corresponding context specified the original noun phrase for each placeholder. The model was instructed to return a JSON object mapping every placeholder to exactly one English pronoun. See the box below.

\begingroup
\setlist{nosep}
\newpage
\section{Comprehensibility of the Discourse-disrupted Texts} \label{app:comp_questions}

To prevent data contamination from the publicly available Natural Stories questions,\footnote{\url{https://github.com/languageMIT/naturalstories/blob/master/naturalstories_RTS/Comp_questions_finalized.txt}} we prompted Gemini-3.6-Thinking to generate new comprehension questions in the same style.
We then prompted DeepSeek-V4-Flash with three versions of the text: unmodified original, \baseline, and \linking, along with the corresponding questions. The task was framed as a forced choice between options (a) and (b). While the correct answer below is always (a), option order was randomized per prompt. 

\subsection*{Story 1}
\begin{enumerate}
    \item What surrounds the valley where the city of Bradford is located?
    \begin{enumerate}
        \item moors as high as mountains
        \item dense oak forests
    \end{enumerate}

    \item At what time of day did the fearsome boar prefer to come out?
    \begin{enumerate}
        \item in the middle of the day
        \item in the dead of night
    \end{enumerate}

    \item Why did the first huntsman decide to cut out the tongue instead of bringing the boar's head?
    \begin{enumerate}
        \item the head was too heavy for him to carry
        \item he wanted to save the head for a feast
    \end{enumerate}

    \item Where did the first huntsman put the boar's tongue after cutting it out?
    \begin{enumerate}
        \item in his hunting pouch
        \item in his coat pocket
    \end{enumerate}

    \item What defect did the Lord of the Manor immediately notice when inspecting the boar's head?
    \begin{enumerate}
        \item the tongue was missing
        \item one of the tusks was broken
    \end{enumerate}

    \item According to the ending of the story, whose fame was ultimately more lasting?
    \begin{enumerate}
        \item the fearsome Bradford Boar's
        \item the victorious huntsman's
    \end{enumerate}
\end{enumerate}

\subsection*{Story 2}
\begin{enumerate}
    \item What did Aqua wish to sail on when he grew tired of playing in the ocean?
    \begin{enumerate}
        \item a white cloud
        \item a sea gull's wing
    \end{enumerate}

    \item Into what form did the sun change the water drops so they were light enough to carry into the sky?
    \begin{enumerate}
        \item fine mist or vapor
        \item soft snowflakes
    \end{enumerate}

    \item Who gathered the frightened water drops into a heavy gray cloud to send them back down to earth?
    \begin{enumerate}
        \item the wind
        \item the sun
    \end{enumerate}

    \item What turned the tremendous wheel inside the flour mill?
    \begin{enumerate}
        \item the tiny water drops
        \item a giant steam engine
    \end{enumerate}

    \item What did Aqua use as horses while playing in the peaceful pond?
    \begin{enumerate}
        \item frogs
        \item ducks
    \end{enumerate}

    \item Where was Aqua at the very end of his journey?
    \begin{enumerate}
        \item back in the ocean
        \item in a deep forest lake
    \end{enumerate}
\end{enumerate}

\subsection*{Story 3}
\begin{enumerate}
    \item Why did the boy take one of the girl's lost slippers?
    \begin{enumerate}
        \item to use it as a home for his pet mouse
        \item to sell it to a shoemaker
    \end{enumerate}

    \item Where did the girl sit down to cover herself when she could not go home?
    \begin{enumerate}
        \item in a corner behind the little bakery
        \item inside a wooden shed
    \end{enumerate}

    \item What did the girl see when she lit her first match?
    \begin{enumerate}
        \item a large iron stove with brass feet
        \item a grand banquet with roast goose
    \end{enumerate}

    \item What did the girl do during her vision of the forest when the second match was lit?
    \begin{enumerate}
        \item dug up food hidden by a squirrel
        \item picked a bouquet of white lilies
    \end{enumerate}

    \item According to her grandmother, what happens when a star falls from the sky?
    \begin{enumerate}
        \item a soul leaves this earth
        \item a child is born
    \end{enumerate}

    \item What did the girl do to keep her grandmother from disappearing when the matches burned out?
    \begin{enumerate}
        \item lit the whole bundle of matches
        \item sang a lullaby
    \end{enumerate}
\end{enumerate}

\subsection*{Story 4}
\begin{enumerate}
    \item Where did the little nameless bird hide right before the flying competition began?
    \begin{enumerate}
        \item on the eagle's back
        \item under the hawk's wing
    \end{enumerate}

    \item Why did the other birds refuse to accept the little bird after he flew higher than the eagle?
    \begin{enumerate}
        \item he broke the rules by cheating
        \item he was too weak to protect them
    \end{enumerate}

    \item What was the second contest proposed to determine who would be king?
    \begin{enumerate}
        \item going the deepest into the earth
        \item finding the largest piece of food
    \end{enumerate}

    \item What hiding place did the little bird creep into for the second contest?
    \begin{enumerate}
        \item a pitch dark mouse's hole
        \item a hollow oak tree
    \end{enumerate}

    \item How did the little bird manage to escape from the owl?
    \begin{enumerate}
        \item the owl fell fast asleep with both eyes shut
        \item the owl left to go hunting
    \end{enumerate}

    \item What do the other birds mockingly call the nameless bird at the end of the story?
    \begin{enumerate}
        \item the hedge-king
        \item the trickster-king
    \end{enumerate}
\end{enumerate}

\subsection*{Story 5}
\begin{enumerate}
    \item Who explained to the narrator what made Elvis Presley so special?
    \begin{enumerate}
        \item Eugene Correthers
        \item the orphanage matron
    \end{enumerate}

    \item What city were the boys taken to for new shoes and haircuts?
    \begin{enumerate}
        \item Florida
        \item Tennessee
    \end{enumerate}

    \item What color did the bones in the boy's feet look through the x-ray machine at the shoe store?
    \begin{enumerate}
        \item green
        \item blue
    \end{enumerate}

    \item Who did the matron speak to after the young barber initially shook his head no?
    \begin{enumerate}
        \item a little man in a squeaky office chair
        \item the shop owner waiting outside
    \end{enumerate}

    \item What did the barber offer the boy after giving him a buzz cut?
    \begin{enumerate}
        \item a nickel for a candy bar
        \item a quarter for a soda
    \end{enumerate}

    \item What question did the boy ask the barber outside at the end of the story?
    \begin{enumerate}
        \item if Elvis Presley has green bones
        \item if Elvis ever got a buzz cut
    \end{enumerate}
\end{enumerate}

\subsection*{Story 6}
\begin{enumerate}
    \item What did Abby draw in art class using two pieces of expensive paper?
    \begin{enumerate}
        \item an elephant
        \item a lion
    \end{enumerate}

    \item What tool did Abby's mother use to clean out the fish tank?
    \begin{enumerate}
        \item a special vacuum
        \item a water filter
    \end{enumerate}

    \item What item did Abby retrieve to help search the tank for Mr.~Sticky?
    \begin{enumerate}
        \item a magnifying glass
        \item a flashlight
    \end{enumerate}

    \item What did Abby briefly mistake for the water snail while searching?
    \begin{enumerate}
        \item a large speck of dust
        \item a small pebble
    \end{enumerate}

    \item Where in the fish tank was Mr.~Sticky finally discovered hiding?
    \begin{enumerate}
        \item in a curve of the archway
        \item inside a castle tower
    \end{enumerate}

    \item What surprise addition did Abby and her mother find right next to Mr.~Sticky?
    \begin{enumerate}
        \item another water snail
        \item a plastic treasure chest
    \end{enumerate}
\end{enumerate}

\subsection*{Story 7}
\begin{enumerate}
    \item How did Lucy respond when her sister asked if she had reported the text messages?
    \begin{enumerate}
        \item she lied and said she had
        \item she admitted she was afraid to
    \end{enumerate}

    \item Whose worn, oversized coat was Lucy wearing to protect herself from the rain?
    \begin{enumerate}
        \item her sister Jill's
        \item her mother's
    \end{enumerate}

    \item What animal was Lucy compared to in the very first abusive text message she received?
    \begin{enumerate}
        \item a giraffe
        \item a flamingo
    \end{enumerate}

    \item In what position was Lucy scheduled to perform in the talent show?
    \begin{enumerate}
        \item first
        \item last
    \end{enumerate}

    \item What encouraging gesture did Jill give Lucy from the wings right before the curtains opened?
    \begin{enumerate}
        \item a thumbs up
        \item a wave
    \end{enumerate}

    \item Who was cheering the loudest in the audience after Lucy finished singing her song?
    \begin{enumerate}
        \item her mother
        \item her school teacher
    \end{enumerate}
\end{enumerate}

\subsection*{Story 8}
\begin{enumerate}
    \item What classified program did the U.S.\ military attribute the recovered debris to?
    \begin{enumerate}
        \item Project Mogul
        \item Project Echo
    \end{enumerate}

    \item What did the Roswell Army Air Field's initial press release on July 9, 1947, claim personnel had recovered?
    \begin{enumerate}
        \item a crashed flying disc
        \item a fallen weather balloon
    \end{enumerate}

    \item Who did Stanton Friedman interview in 1978 about transporting the debris to Fort Worth?
    \begin{enumerate}
        \item Jesse Marcel
        \item Karl Pflock
    \end{enumerate}

    \item What major flaw did critics point out regarding all the witness accounts collected?
    \begin{enumerate}
        \item they came a minimum of thirty-one years after the event
        \item they were contradicted by radar logs
    \end{enumerate}

    \item According to the Air Force's 1997 report, what contributed to the reports of recovered alien bodies?
    \begin{enumerate}
        \item memories of military accidents and anthropomorphic test dummies
        \item secret biological warfare experiments
    \end{enumerate}

    \item During which holiday weekend does the City of Roswell host its annual Roswell UFO Festival?
    \begin{enumerate}
        \item July Fourth weekend
        \item Memorial Day weekend
    \end{enumerate}
\end{enumerate}

\subsection*{Story 9}
\begin{enumerate}
    \item From which empire was the tulip introduced to Europe in the mid-sixteenth century?
    \begin{enumerate}
        \item the Ottoman Empire
        \item the Persian Empire
    \end{enumerate}

    \item What was responsible for the vivid, multicolored patterns on the most sought-after tulip petals?
    \begin{enumerate}
        \item a tulip-specific virus
        \item a rare soil mineral
    \end{enumerate}

    \item How long does it take for a tulip seed to form a flowering bulb?
    \begin{enumerate}
        \item seven to twelve years
        \item one to two years
    \end{enumerate}

    \item Where did merchants and traders gather to buy and sell tulip futures contracts?
    \begin{enumerate}
        \item in taverns
        \item in royal courtyards
    \end{enumerate}

    \item Who wrote the 1841 book \textit{Extraordinary Popular Delusions and the Madness of Crowds}?
    \begin{enumerate}
        \item Charles Mackay
        \item Charles de l'Ecluse
    \end{enumerate}

    \item Why did Dutch courts refuse to enforce payments on tulip contracts after the market crashed?
    \begin{enumerate}
        \item they regarded the debts as contracted through gambling
        \item the government had outlawed flower sales
    \end{enumerate}
\end{enumerate}

\subsection*{Story 10}
\begin{enumerate}
    \item In what person did Dr.~Georges Gilles de la Tourette first describe the disorder?
    \begin{enumerate}
        \item an eighty-six-year-old French noblewoman
        \item a ten-year-old English schoolboy
    \end{enumerate}

    \item What condition do parents often mistake sniffing tics for before a formal diagnosis?
    \begin{enumerate}
        \item seasonal allergies
        \item a chronic sinus infection
    \end{enumerate}

    \item What is the term for complex vocal tics that involve uttering swear words?
    \begin{enumerate}
        \item coprolalia
        \item echolalia
    \end{enumerate}

    \item For at least how long must a patient display both motor and vocal tics to receive a formal diagnosis of Tourette's?
    \begin{enumerate}
        \item at least one year
        \item at least six months
    \end{enumerate}

    \item What is the main purpose of using neuroimaging studies like MRIs or CT scans during evaluation?
    \begin{enumerate}
        \item to rule out other conditions
        \item to measure the severity of the tics
    \end{enumerate}

    \item Females with a genetic predisposition for the disorder are more likely than males to exhibit which type of symptoms?
    \begin{enumerate}
        \item obsessive-compulsive symptoms
        \item severe motor tics
    \end{enumerate}
\end{enumerate}

\end{document}